\documentclass[preprint]{vgtc}               %

\ifpdf%
  \pdfoutput=1\relax                   %
  \usepackage{graphicx}                %
  \DeclareGraphicsExtensions{.pdf,.png,.jpg,.jpeg} %
\else%
  \usepackage{graphicx}                %
  \DeclareGraphicsExtensions{.eps}     %
\fi%

\graphicspath{{figures/}{pictures/}{images/}{./}} %

\usepackage{microtype}                 %
\PassOptionsToPackage{warn}{textcomp}  %
\usepackage{textcomp}                  %
\usepackage{mathptmx}                  %
\usepackage{times}                     %
\usepackage{cite}                      %

\usepackage[utf8]{inputenc}
\usepackage[T1]{fontenc}

\usepackage{mathtools}
\usepackage{amssymb, amsfonts}
\usepackage{bm} %

\usepackage{wrapfig}
\usepackage{dblfloatfix}
\usepackage{placeins}

\usepackage{subcaption}
\usepackage{pgfplots}
\pgfplotsset{compat=1.16} 
\usepackage{tikz}
\usepackage{siunitx}
\usepackage{enumitem}

\usepackage[acronym,nomain]{glossaries}

\newacronym{gis}{GIS}{geographic information system}
\newacronym{qf}{QF}{quadratic form}
\newacronym{emd}{EMD}{Earth Mover's Distance}
\newacronym{lisa}{LISA}{Local Indicator of Spatial Association}
\newacronym{pc}{PC}{point cloud}

\newcommand{\imgHighDim}{\ensuremath{f}}
\newcommand{\imgOneDim}[1][c]{\ensuremath{f_{#1}}} 
\newcommand{\imgX}{\ensuremath{X}}

\newcommand{\imgY}{\ensuremath{Y}}

\newcommand{\imgSizeXY}{\ensuremath{|X \times Y|}}
\newcommand{\imgSizeTot}{\ensuremath{n}}
\newcommand{\neighborhodSizeTot}{\ensuremath{M}}
\newcommand{\numAttrDimensions}{\ensuremath{C}}
\newcommand{\imgPos}[1][i]{\ensuremath{\mathbf{#1}}}
\newcommand{\numSpNeigh}{\ensuremath{\eta}}
\newcommand{\spatialNeighborhood}{\ensuremath{\mathcal{N}^{S,\numSpNeigh}}}
\newcommand{\attrNeighborhood}{\ensuremath{\mathcal{N}^{C,k}}}

\newcommand{\dataPointHighDim}{\ensuremath{\mathbf{g}}}
\newcommand{\dataPointScalar}{\ensuremath{g}} %
\newcommand{\featureMat}{\ensuremath{\bm{T}}}

\newcommand{\featureExtract}[1][\numSpNeigh]{\ensuremath{\mathcal{T}_{#1}}} %

\newcommand{\standardDist}{\ensuremath{d}}
\newcommand{\spatialDist}[1][]{\ensuremath{d_{s}^{#1}}}

\newcommand{\weightsvec}{\ensuremath{\mathbf{w}}}
\newcommand{\weight}{\ensuremath{w}}

\newcommand{\histogram}{\ensuremath{\bm{h}}}
\newcommand{\histogramBin}{\ensuremath{h}}
\newcommand{\covmat}{\ensuremath{\bm{\Sigma}}}
\newcommand{\cov}{\ensuremath{\sigma}}
\newcommand{\meanvec}{\ensuremath{\mathbf{\mu}}}
\newcommand{\mean}{\ensuremath{\mu}}

\definecolor{plotcolorblue}{HTML}{00a2ff}
\definecolor{plotcolorgreen}{HTML}{61d836}
\definecolor{plotcolorred}{HTML}{ff644e}
\definecolor{plotcoloryellow}{HTML}{f8ba00}

\definecolor{plotcolorblue1}{HTML}{004d80}
\definecolor{plotcolorblue2}{HTML}{0076ba}
\definecolor{plotcolorblue3}{HTML}{56c1ff}
\definecolor{plotcolorpink1}{HTML}{970e53}
\definecolor{plotcolorpink2}{HTML}{d41876}
\definecolor{plotcolorpink3}{HTML}{ff42a1}

\makeatletter
\renewcommand*{\@fnsymbol}[1]{\ensuremath{\ifcase#1\or *\or \dagger\or \ddagger\or
   \mathsection\or \mathparagraph\or \|\or **\or \dagger\dagger
   \or \ddagger\ddagger \else\@ctrerr\fi}}
\makeatother

\makeatletter
\newcommand\Autoref[1]{\@first@ref#1,@}
\def\@throw@dot#1.#2@{#1}%
\def\@set@refname#1{%
    \edef\@tmp{\getrefbykeydefault{#1}{anchor}{}}%
    \xdef\@tmp{\expandafter\@throw@dot\@tmp.@}%
    \ltx@IfUndefined{\@tmp autorefnameplural}%
         {\def\@refname{\@nameuse{\@tmp autorefname}s}}%
         {\def\@refname{\@nameuse{\@tmp autorefnameplural}}}%
}
\def\@first@ref#1,#2{%
  \ifx#2@\autoref{#1}\let\@nextref\@gobble%
  \else%
    \@set@refname{#1}%
    \@refname~\ref{#1}%
    \let\@nextref\@next@ref%
  \fi%
  \@nextref#2%
}
\def\@next@ref#1,#2{%
   \ifx#2@ and~\ref{#1}\let\@nextref\@gobble%
   \else, \ref{#1}%
   \fi%
   \@nextref#2%
}

\makeatother

\usepackage{multicol}
\usepackage{setspace}

\usepackage{placeins}

\usepackage{xparse}
\usepackage[user,hyperref]{zref}
\makeatletter
\providecommand{\@currentshorttitle}{}
\zref@newprop{shorttitle}{\@currentshorttitle}
\zref@addprop{main}{shorttitle}

\NewDocumentCommand{\labelshort}{om}{%
  \begingroup
  \IfValueT{#1}{%
    \renewcommand{\@currentshorttitle}{#1}%
    \zlabel{#2}%
  }%
  \endgroup
  \label{#2}%
}

\NewDocumentCommand{\nameshortref}{O{}m}{%
  \zref@ifrefundefined{#2}{%
  }{%
    \hyperlink{\zref@extract{#2}{anchor}}{#1\zref@extract{#2}{shorttitle}}%
  }%
}

\usepackage[resetlabels]{multibib}
\newcites{supplement}{Supplemental references}

\DeclareMathOperator*{\kargmin}{\mathit{k}-arg\,min}

\usepackage[compact]{titlesec}
\titlespacing*{\section}{0pt}{1.4ex plus .0ex minus -.4ex}{0.5ex plus .2ex minus -.4ex}
\titlespacing*{\subsection}{0pt}{1.2ex plus .0ex minus -.2ex}{0.3ex plus .2ex minus -.4ex}

\onlineid{4701}

\vgtccategory{Technique}  %

\vgtcinsertpkg

\title{Incorporating Texture Information into\\ Dimensionality Reduction for High-Dimensional Images}

\author{A. Vieth\thanks{e-mail: \{A.Vieth~$|$~E.Eisemann~$|$~T.Hollt-1\} @tudelft.nl}\\ %
      \parbox{0.9in}{\scriptsize \centering TU Delft \\ The Netherlands} %
\and A. Vilanova\thanks{e-mail: A.Vilanova@tue.nl}\\ %
      \parbox{0.9in}{\scriptsize \centering TU Eindhoven \\ The Netherlands} %
\and B. Lelieveldt\thanks{e-mail: B.P.F.Lelieveldt@lumc.nl}\\ %
      \parbox{1.6in}{\scriptsize \centering Leiden University Medical Center \\ The Netherlands} %
\and E. Eisemann$^*$ \\ %
      \parbox{0.9in}{\scriptsize \centering TU Delft \\ The Netherlands} %
\and T. Höllt$^*$ \\ %
      \parbox{0.9in}{\scriptsize \centering TU Delft \\ The Netherlands}}

\abstract{High-dimensional imaging
is becoming increasingly relevant in many fields from astronomy and cultural heritage to systems biology.
Visual exploration of such high-dimensional data is commonly facilitated by dimensionality reduction.
However, common dimensionality reduction methods do not include spatial information present in images, such as local texture features, into the construction of low-dimensional embeddings.
Consequently, exploration of such data is typically split into a step focusing on the attribute space followed by a step focusing on spatial information, or vice versa.
In this paper, we present a method for incorporating spatial neighborhood information into distance-based dimensionality reduction methods, such as t-Distributed Stochastic Neighbor Embedding (t-SNE).
We achieve this by modifying the distance measure between high-dimensional attribute vectors associated with each pixel such that it takes the pixel's spatial neighborhood into account.
Based on a classification of different methods for comparing image patches, we explore a number of different approaches.
We compare these approaches from a theoretical and experimental point of view. 
Finally, we illustrate the value of the proposed methods by qualitative and quantitative evaluation on synthetic data and two real-world use cases.}

\CCScatlist{
  \CCScatTwelve{Mathematics of computing}{Dimensionality reduction}{}{};
  \CCScatTwelve{Human-centered computing}{Visualization techniques}{}{};
  \CCScatTwelve{Human-centered computing}{Visual analytics}{}{};
}

\begin{document}

\firstsection{Introduction}

\maketitle

\begin{figure}[t!]
    \begin{center}
        \includegraphics[width=\linewidth]{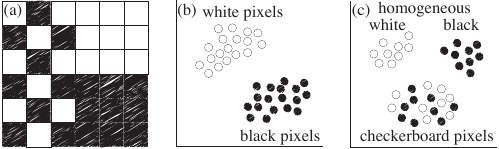}%
        \phantomsubcaption\ignorespaces\label{fig:introillustration_image}%
        \phantomsubcaption\ignorespaces\label{fig:introillustration_standard}%
        \phantomsubcaption\ignorespaces\label{fig:introillustration_enhanced}%
        \vspace{-3mm} %
    \end{center}
    \caption{\textbf{Texture-aware dimensionality reduction.} An image (a) with black and white pixels forms multiple textures. Standard distance-based dimensionality reduction produces one cluster of black and one cluster of white pixels (b), a texture-aware version should create clusters for the different textures (c).}
    \label{fig:introillustration}
\end{figure}

High-dimensional data is commonly acquired and analyzed in various application domains, from systems biology~\cite{bib:hollt:2019b} to insurance fraud detection~\cite{bib:leite:2018}.
Typically, high-dimensional data are tabular data with many columns (or attributes), corresponding to the dimensionality per item but there are no connections between items.
Dimensionality reduction techniques like t-distributed Stochastic Neighbor Embedding (t-SNE)~\cite{VanderMaaten2008} or Uniform Manifold Approximation and Projection (UMAP)~\cite{mcinnes2018umap} are well-established tools used for exploratory visual analysis of such high-dimensional data~\cite{bib:sedlmair:2013}.
Advances in imaging techniques have introduced an increasing number of imaging data modalities producing high-dimensional images (every pixel represents a high-dimensional attribute-vector).
Current state-of-the-art dimensionality reduction methods are commonly used for the explorative analysis of such imaging data, for example in cultural heritage~\cite{bib:alfeld:2018},
biology~\cite{bib:devries:2020}, or geospatial applications~\cite{bib:dabrowska:2020}.
However, they rely only on attribute data of pixels and do not take  additional spatial information, such as texture, present in such imaging data into account.
Thus, in the resulting low-dimensional embeddings, the pixels are only arranged according to their individual attributes~(\autoref{fig:introillustration_standard}), but do not provide any insight into texture, neighborhoods or other spatial relations common in image analysis.

The spatial configuration is, however, commonly of interest when analyzing high-dimensional image data.
For example, taking spatial neighborhood information into account, in addition to high-dimensional attributes, has led to new discoveries in single-cell biology~\cite{bib:2021_acta_neuropathologica_communications}.
Typical approaches to combine high-dimensional attributes and spatial information, however, rely on a two-stage process:
first, high-dimensional attributes are aggregated, for example to classify pixels, then standard image analysis is performed on the aggregate images.
For example, Abdelmoula et al.~\cite{Abdelmoula2016} aggregate the original attribute space of  high-dimensional imaging data to a three-dimensional space using t-SNE.
They use the embedding as a colormap and perform segmentation on the re-colored image.
Decoupling the high-dimensional and spatial analysis in such a way has several downsides:
Most importantly, boundaries between clusters in an embedding are often not well defined, and as such classification is ambiguous and has a level of arbitrariness.
Issues with inaccurate classification might appear undetected and lead to wrong conclusions. 
Furthermore, if problems with the classification become apparent in the spatial analysis, one has to go back to the high-dimensional analysis and potentially loses all progress in the spatial analysis.
Moreover, the necessary aggregation in the  first step limits what is discoverable in the spatial analysis step.
Therefore, we deem the integration of spatial information directly into the dimensionality reduction desirable for exploratory analysis.

We present an approach to integrate spatial information directly into the dimensionality reduction process with the goal to combine attribute and spatial information in a single embedding~(\autoref{fig:introillustration_enhanced}).
Specifically, we propose to adapt the similarity computation, used in distance-based dimensionality reduction, such as t-SNE, UMAP, or Multi-dimensional scaling (MDS), to incorporate different spatial neighborhood features.
We exemplarily present different similarity computation methods for such neighborhood comparisons by extending an existing classification~\cite{Zitova2003} to high-dimensional images.

The main contributions of this paper are: %
\begin{itemize}[topsep=2pt,itemsep=2pt,partopsep=2pt, parsep=2pt, leftmargin=0.35cm]
    \item incorporating texture information into distance-based dimensionality reduction for exploratory analysis of high-dimensional images through distance measures including image neighborhoods. %
	\item the exemplary extension of t-SNE using different classes of neighborhood distance measures and their analysis.%
\end{itemize}

\section{Related Work}  \label{sec:relatedWork}
Dimensionality reduction and high-dimensional images are well-researched topics with a large body of work.
In the following, we aim to report the work most relevant to our contribution.

\subsection{Explorative analysis of high-dimensional images} \label{sec:relatedWork:VisualAnalytics}
There is a multitude of approaches for visual exploration of high-dimensional data~\cite{Liu2017}.
A challenge when facing data that is additionally spatially resolved, is to effectively visualize spatial and attribute characteristics in an integrated fashion.
MulteeSum~\cite{Meyer2010} compares spatio-temporal gene expression data from fruit fly embryos by segmenting cells in the image and providing multiple attribute summaries per cell.  %
Another approach is to characterize the data attributes in terms of specific features and represent them as glyphs at their respective regions in space~\cite{Eichner2013}.
For multivariate volume data, high-dimensional transfer functions have been employed in combination with standard volume rendering techniques~\cite{Zhou2013}.

Often, high-dimensional image data is visualized indirectly by first extracting features that capture interesting data characteristics and then displaying those~\cite{Kehrer2013, Fuchs2009}.
One such feature is texture.
We follow Haidekker's definition of texture as "any systematic local variation of the image values"~\cite{HaidekkerM2010} since it emphasises that texture encodes spatially local relationships of pixel values.
Texture feature extraction is a broad, well-established field~\cite{Humeau-Heurtier2019} but the extension of single-channel texture features to multi-channel images is not trivial.
Typically, single-channel texture features are extracted for each channel and concatenated to a feature vector. 
Multi-channel texture features such as color co-occurrence matrices proposed by Palm~\cite{Palm2004} are less common and
engineering such features is an ongoing process in the image processing community~\cite{Singh2018}.

Dimensionality reduction has been used to \emph{create} texture features~\cite{Lefebvre2006} and \emph{explore} texture databases~\cite{Luo2021} but such approaches are generally out of the scope of this work.

\vspace{2mm}
\hspace{-4.5mm}
\begin{minipage}{9.0cm}
\subsection{
    \hspace{-4mm}
        Dimensionality reduction for high-dimensional images
} \label{sec:relatedWork:DRofHDImages}
\end{minipage}
Non-linear dimensionality reduction techniques such as t-SNE~\cite{VanderMaaten2008} and UMAP~\cite{mcinnes2018umap} have become popular techniques to visualize and explore high-dimensional data.
These techniques have been applied to high-dimensional imaging data without considering texture information.
Abdelmoula et al.~\cite{Abdelmoula2016} use t-SNE in a segmentation pipeline for high-dimensional images. 
They embed pixels according to their high-dimensional attribute values to a three-dimensional space, followed by coloring the pixels by using the 3D embedding coordinates as coordinates in the $\textrm{L}^{*}\textrm{a}^{*}$b color space.
The resulting color images are then used to aid the segmentation with the goal of identifying tissue segments with similar properties, according to the original attribute space.
Recently, Evers et al.~\cite{Evers2021} followed a similar approach to identify regional correlations in spatio-temporal weather ensemble simulations with the main difference of using MDS instead of t-SNE.
Others combine dimensionality reduction with segmented image data.
Facetto~\cite{Krueger2019} combines un- and semi-supervised learning to aid in the visual analysis of high-dimensional imaging data in the field of structural biology.
After segmenting cells and aggregating their corresponding attributes to features, they use UMAP to display the cells according to their similarities.
ImaCytE~\cite{Somarakis2019} is a visual analysis tool for similar data that focuses on the analysis of cell neighborhoods.
Again, cells are segmented and the attributes of pixels within the cells aggregated. 
Cells are laid out according to their similarity in the attribute space using t-SNE and the resulting information is used to analyze spatial neighborhoods. %
All of these applications and tools make use of standard dimensionality reduction methods that do not incorporate spatial information and instead follow a two-step approach using the results of either dimensionality reduction or spatial analysis as input to the other. 
Instead, we directly include spatial information in the dimensionality reduction process to reduce the number of steps and potential points of failure in interactive analysis.

In the analysis of hyperspectral images, dimensionality reduction is an important step for pixel classification.
A common approach to include spatial information relies on computing the first couple of principal components of the high-dimensional data and then continuing with classic image processing methods that work on scalar or color data to extract spatial neighborhood information~\cite{Fauvel2013, Ghamisi2017}.
For example, morphological image processing techniques are used to capture spatial structure in high-dimensional images~\cite{DallaMura2010}.
Spatial-spectral local discriminant projection~\cite{Huang2019a} takes a more direct way of combining spatial and spectral information into dimensionality reduction by incorporating a weighting factor into the neighborhood preserving embedding that represents the spectral similarity between spatially neighboring pixels. 
However, this and similar hyperspectral image analysis methods~\cite{Lu2014} rely on training with ground truth data, which is typically not available in exploratory data analysis.
Recently, Halladin-Dąbrowska et al.~\cite{bib:dabrowska:2020} proposed a workflow using t-SNE for cleaning ground truth data, however, they do not include spatial neighborhood information in their embeddings.

A straightforward way to inform dimensionality reduction techniques of images about their spatial domain is to consider each data point's spatial location in the point similarity measure used during the embedding. 
Spherical SNE~\cite{Lunga2013} devises a similarity function between data points in the style of bilateral filtering that weights attribute distance with pixel location distance.
This approach, however, does not capture the similarity of the local structure around the compared points, which we aim for.

\vspace{2mm}
\subsection{Multivariate graph visualization and embedding} \label{sec:relatedWork:networks}
Graph-based techniques are commonly applied to pattern recognition and computer vision problems on imaging data~\cite{Conte2007}.
For that, images are interpreted as graphs, where each pixel is interpreted as a node and neighboring pixels are connected by a link. 
Several techniques for graph drawing aim to incorporate \textit{network structure} and \textit{node attributes} exist~\cite{Kerren2014} that might be helpful for our goals.

GraphTPP~\cite{Gibson2017} focuses on a visual combination of node attributes and connections in a 2D graph layout. 
First, principal component analysis (PCA) is applied to  the data using only the attributes.
Then links between nodes are overlaid on the resulting scatterplot. 
The user can then manually reposition points according to their interpretation, and compute a new linear projection that best fits the modified layout.
GraphTSNE~\cite{Leow2019} aims to preserve graph connectivity and node attribute similarity.
It does so by training a graph convolutional network on a modified t-SNE loss that combines the squared euclidean distance between node attributes and the shortest-path distance between the nodes on the graph. 
Their design seeks to position two points close in the embedding either when their attributes are similar or they are connected by an edge.
Similarly, MVN-Reduce~\cite{eurovisshort.20171126} defines a distance measure between two nodes as the sum of a node's attribute distance and their weighted shortest path distance on the graph.
The resulting distances are used as input to distance-based dimensionality reduction methods like MDS.
The Heterogeneous Network Embedding (HNE) framework~\cite{Chang2015} aims to create embeddings that position data points with links closer and those without further away from each other. 
Therein, a neural network is trained with a loss function that builds on a similarity term between point attributes that is weighted depending on their respective node linkage.

Applied to an image, transformed into a graph, all the aforementioned approaches essentially combine the pixel location distance (geodesic distances on the graph) with the attribute distance, not dissimilar to what is described for the Spherical SNE~\cite{Lunga2013} (\autoref{sec:relatedWork:DRofHDImages}).
In contrast, the goal of our proposed approach is to compare local texture structures rather than absolute distances. 
Two pixels are compared by taking into account the structure of the high-dimensional values in the spatial neighborhoods of the two pixels.

\section{Background}  \label{sec:background}
We will first define high-dimensional image data and spatial neighborhoods more rigorously in~\Autoref{sec:background:data, sec:background:neighborhood} respectively. 
In \autoref{sec:background:t-SNE}, we briefly introduce t-SNE as an example of distance-based dimensionality reduction methods. %

\begin{figure}[t!]
    \begin{center}
        \includegraphics[width=\linewidth]{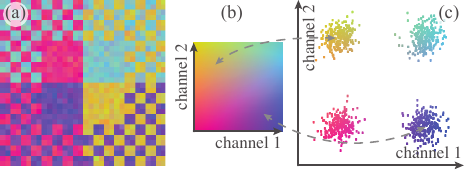}
        \phantomsubcaption\ignorespaces\label{fig:ArtStructure_image}   
        \phantomsubcaption\ignorespaces\label{fig:ArtStructure_colormap}
        \phantomsubcaption\ignorespaces\label{fig:ArtStructure_pc}
        \vspace{-5mm}
    \end{center}
    \vspace{-3mm}
    \caption{\textbf{Synthetic test image dataset} with $X \times Y = 32 \times 32$ and $C=2$. The distribution in attribute space (c) reveals four groups. We use a 2D colormap (b) to color pixels in image space (a) according to their attribute information.\vspace{-5mm}}
    \label{fig:ArtStructure}

\end{figure}

\subsection{High-dimensional image data}  \label{sec:background:data}
While, in general, dimensionality reduction methods can be applied on other high-dimensional data sets as well, we focus on high-dimensional, digital 2D images.
Each pixel is an image element at a unique location with an associated attribute vector.
We can formalize such an image as a discrete function ${\imgHighDim : \imgX \times \imgY \to \mathbb{R}^\numAttrDimensions}$ from the spatial domain $\imgX \times \imgY \subset \mathbf{N}^2$ to the attribute range ${\mathbb{R}^\numAttrDimensions}$.
\imgX\ and \imgY\ are the sets of pixel coordinates that span the image domain along its two dimensions while $\numAttrDimensions$ is the number of attributes, or image channels.
A pixel is indexed with ${\imgPos = (x_i, y_i)}$ where ${1 \leq i \leq \imgSizeTot}$ and ${n = \imgSizeXY}$. 
Now, ${\imgHighDim(\imgPos) =\dataPointHighDim_{\imgPos}}$ yields the pixel's high-dimensional attribute vector at location~$\imgPos$ with ${\dataPointHighDim_{\imgPos} = [\dataPointScalar_{\imgPos 1}, \dots, \dataPointScalar_{\imgPos \numAttrDimensions}]}$. 

We can also interpret such a high-dimensional image as a combination of several scalar images, each representing one channel of the high-dimensional image.
Therefore, we refer to the~$c$-th channel of the image with $\imgOneDim$, where ${1 \leq c \leq \numAttrDimensions}$, such that ${\imgOneDim = [\dataPointScalar_{\mathbf{1} c}, \dots, \dataPointScalar_{\mathbf{n} c} ]}$ denotes the values of the~$c$-th channel for all pixels in the image.

\autoref{fig:ArtStructure} shows a synthetic toy-example of a `high-dimensional' image with two attribute channels (i.e., $\numAttrDimensions=2$). %
The spatial layout is displayed in \autoref{fig:ArtStructure_image} with each pixel color coded according its two attribute values using a 2D colormap (\autoref{fig:ArtStructure_colormap}). 
\autoref{fig:ArtStructure_pc} shows a scatterplot of all attribute values.
Four groups are clearly distinguishable based on the attributes.
In image space the four groups form eight visually distinct regions, four group-homogeneous and four consisting of checkered patches.
In \autoref{sec:synthetic}, we use this image to showcase the characteristics of our proposed approaches and compare it to a standard t-SNE embedding.

\subsection{Neighborhood definition}\label{sec:background:neighborhood}
The notion of neighborhood is twofold for high-dimensional image data: 
one can distinguish between neighbors in the spatial and the attribute domain. 
We refer to the \emph{attribute neighborhood}~$\attrNeighborhood_{\imgPos}$ of pixel~$\imgPos$ as the set of indices of the~$k$ pixels with the smallest distances~$\standardDist$ to the attributes of pixel~$\imgPos$:
\begin{equation}
	\attrNeighborhood_{\imgPos} = \kargmin_{\imgPos[j] = (x_j, y_j),\ 1 \leq j \leq \imgSizeTot} \standardDist( \dataPointHighDim_{\imgPos}, \dataPointHighDim_{\imgPos[j]}) ,
	\label{eq:attributeNeigh}
\end{equation}
where $\kargmin$ performs a selection of the $k$ arguments in ${\{ \imgPos[j] = (x_j, y_j) : \ 1 \leq j \leq \imgSizeTot \}}$ that yield the $k$ smallest distances.
Typically, the squared Euclidean distance is chosen as the distance measure $\standardDist(\dataPointHighDim{_{\imgPos}}, \dataPointHighDim_{\imgPos[j]}) = \lVert \dataPointHighDim_{\imgPos} - \dataPointHighDim_{\imgPos[j]} \rVert^2_2$, but other distances like cosine or Hamming distance are popular choices as well.

Next, we define the \emph{spatial neighborhood} $\spatialNeighborhood_{\imgPos}$. We consider a square spatial neighborhood of ${\neighborhodSizeTot = (2 \cdot \numSpNeigh + 1)^2}$ pixels centered at~$\imgPos$.
With~${\imgPos = (x_i, y_i)}$, this results in the set of spatially-neighboring indices%
\begin{equation}
		\spatialNeighborhood_{\imgPos} = \{ (x_i + r,\ y_i + s)\ :\ -\numSpNeigh \leq\ r,\ s \leq\ \numSpNeigh \}.
	\label{eq:spatialNeigh}
\end{equation}
We focus on two-dimensional, rectilinear spatial layouts for the sake of simplicity, but in principle our method is trivially extendable for data resolved in three spatial dimensions, i.e., multivariate volumetric data: %
{$\spatialNeighborhood_{\imgPos} = \{ (x_i + r,\ y_i + s,\ z_i + t)\ :\ -\numSpNeigh \leq\ r,\ s,\ t \leq\ \numSpNeigh \}.$}

\begin{figure*}[ht]
    \centering
    \vspace{-4mm}
    \includegraphics[width=\textwidth]{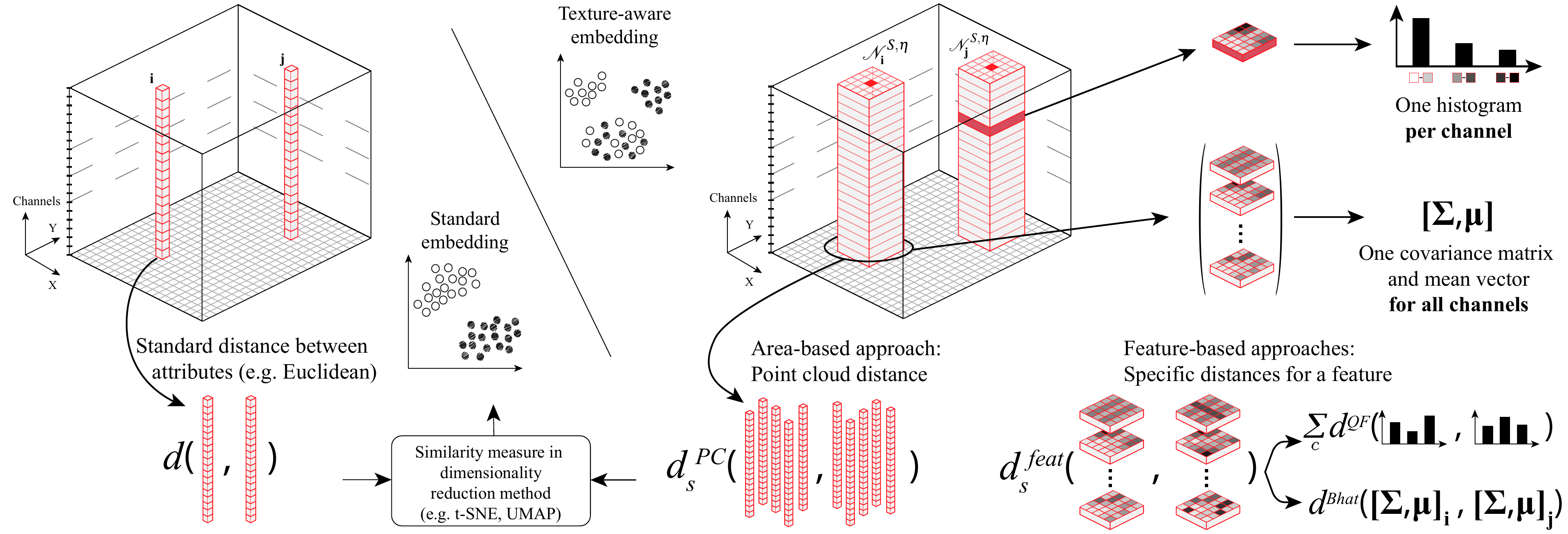}
    \vspace{-6mm}
    \caption{\textbf{Incorporating image texture information into dimensionality reduction} by adapting the distance measure that defines pixel similarity. High-dimensional image data is depicted in a data cube. Standard t-SNE compares pixels based  on their attribute vectors only, e.g., using Euclidean distance (left). We propose to also consider the spatial neighborhood~$\spatialNeighborhood_{\imgPos}$ of the pixels (right) and present different approaches. Feature-based methods (FBM) derive texture features per channel (e.g., local histogram) or across channels (e.g, covariance matrix) and compare those, while area-based method (ABM) (e.g., point cloud distance) compare sets of original attribute vectors directly.} \label{fig:method-overivew}
    \vspace{-4mm}
\end{figure*}

\subsection{t-SNE}  \label{sec:background:t-SNE}
Distance-based dimensionality reduction methods like LargeVis~\cite{Tang2016}, UMAP~\cite{mcinnes2018umap} and t-SNE~\cite{VanderMaaten2008, VanderMaaten2014} all share a similar basic structure.
First, based on a distance measure %
in the attribute space, they construct a neighborhood graph that captures local neighborhoods.
Then, a low dimensional layout is produced, with the aim to represent these neighborhoods as faithfully as possible; this process is guided by optimizing a specific cost function.
We will discuss our framework using the example of t-SNE.
While the same concepts are applicable to all distance-based dimensionality reduction methods, we deem it easiest to follow one specific example.

To create a low dimensional embedding as described above, t\nobreakdash-SNE uses a symmetric joint probability distribution $P$ to describe similarities between high-dimensional points. 
Likewise, a joint probability distribution $Q$ encodes the similarity of the corresponding low-dimensional points in the embedding space.
Starting with a random initialization of $Q$, the embedding points are iteratively moved such that the distribution $Q$ matches $P$ well.
This optimization process is guided by the Kullback-Leibler ($KL$) divergence that measures the divergence of $P$ and $Q$ as cost function $Cost(P,Q)$ 
\begin{equation}
    Cost(P,Q) = KL(P,Q) = \sum_{i}^{\imgSizeTot} \sum_{j,j\neq i}^{\imgSizeTot} p_{\imgPos\imgPos[j]} \ln{\left( \frac{p_{\imgPos\imgPos[j]}}{q_{\imgPos\imgPos[j]}} \right)},
    \label{eq:tSNECost}
\end{equation}
where the probability $p_{\imgPos\imgPos[j]}$ represents the similarity of two high-dimensional data points $\dataPointHighDim_{\imgPos}$ and $\dataPointHighDim_{\imgPos[j]}$, and $q_{\imgPos\imgPos[j]}$ represent the similarity of the two corresponding low-dimensional data points in the embedding.
$p_{\imgPos\imgPos[j]}$ is symmetric and computed as
\begin{equation}
    p_{\imgPos\imgPos[j]} = \frac{p_{\imgPos|\imgPos[j]} + p_{\imgPos[j]|\imgPos}}{2 \imgSizeTot},
    \label{eq:tSNEpij}
\end{equation}
where $p_{\imgPos[j]|\imgPos}$ can be interpreted as the probability that the point $\dataPointHighDim_{\imgPos[j]}$ is in the neighborhood of the point $\dataPointHighDim_{\imgPos}$ in the attribute space.
$p_{\imgPos[j]|\imgPos}$ is calculated using the distance measure $\standardDist(\dataPointHighDim_i, \dataPointHighDim_{\imgPos[j]})$ between the high-dimensional points:
\begin{equation}
    p_{\imgPos[j]|\imgPos} = 
    \begin{cases}
       \frac{\exp{ \left( - d(\dataPointHighDim_{\imgPos}, \dataPointHighDim_{\imgPos[j]}) / (2 \sigma_{\imgPos}^2) \right)} } 
            {\sum_{\imgPos[k] \in \attrNeighborhood_{\imgPos}} \exp{ \left( - d(\dataPointHighDim_{\imgPos}, \dataPointHighDim_{\imgPos[k]}) / (2\sigma_{\imgPos}^2) \right)} } 
                	 	 & \text{if } \imgPos[j] \in \attrNeighborhood_{\imgPos} \\
       \hspace{45pt} 0   & \text{otherwise.}
    \end{cases}
	\label{eq:tSNEpj|i}
\end{equation}

The number of nearest neighbors ${|\attrNeighborhood_{\imgPos}| = 3 \varphi}$ can be steered with a user-defined perplexity~$\varphi$.
The bandwidth $\sigma_i$, in turn, is determined based on the given perplexity value such that
\begin{equation}
    \varphi = 2^{-\sum_{\imgPos[j] \in \attrNeighborhood_{\imgPos}} p_{\imgPos[j]|\imgPos} \log_2 p_{\imgPos[j]|\imgPos}} .
    \label{eq:perplexity}
\end{equation}

\section{Texture-Aware Dimensionality Reduction } \label{sec:method}
To incorporate spatial neighborhood information into low-dimensional embeddings, we propose a set of distance measures that take the spatial neighborhood $\spatialNeighborhood$ of pixels in high-dimensional images into account.
The distance between the attributes of two attribute vectors ${\standardDist(\dataPointHighDim_{\imgPos}, \dataPointHighDim_{\imgPos[j]})}$ 
is thus replaced by a new texture-informed distance ${\spatialDist(\imgPos, \imgPos[j], \imgHighDim, \numSpNeigh})$ and the k nearest neighbors in \autoref{eq:tSNEpj|i} will be based on this new measure as well.
Since we aim to compare the spatial neighborhood of two pixel, it does not suffice to include their image coordinates or spatial distance; rather, it is necessary to involve each pixel's spatial local neighborhood.
Essentially, we are comparing image patches instead of single pixel values.
All other steps of the embedding process remain as they were.

\subsection{Comparing image patches}  \label{sec:method:ABMFBM}
In the following, we will present a number of texture-aware distance measures ${d_s}$. %
As the space of potential measures is vast, we will focus on a few exemplary measures, following the classification of distance measures for image patches introduced by Zitová and Flusser~\cite{Zitova2003}.
In particular, they distinguish image patch comparison into area-based methods (ABM) and feature-based methods (FBM).

ABM and FBM for image patch comparison differ in their approach to compute similarity scores.
ABM directly work with the pixel's attribute values of the two image patches to compare.
They are sometimes called intensity-based instead, which might reflect the immediate usage of the attribute values more aptly.  
In contrast, FBM follow a two-step approach of first computing features for each patch and then comparing those.
ABM can be further categorized into correlation-like methods (for example point cloud distances), Fourier methods, mutual information methods, optimization methods~\cite{Zitova2003}. 
We refer to Goshtasby~\cite{Goshtasby2012} for an extensive overview and discussion of area-based similarity and dissimilarity measures.
Likewise, there exists a rich body of literature
discussing image patch descriptors and appropriate feature matching methods used in FBM~\cite{Kapoor2021, Leng2019}.
To note, most FBM extract various salient features per image with the goal of matching the whole images.
Such methods do not necessarily produce one feature value per pixel which is desired for computing the pairwise distance between pixels as in our case.
More extensive and general discussions of ABM and FBM can be found in~\cite{Zitova2003, Ma2020}.
Here, we will present how to structurally extend image patch comparison to high-dimensional images, and will showcase t-SNE with exemplary methods for each category.

\subsection{Application to high-dimensional images}  \label{sec:method:ABMFBMinHighDim}
Classically, both FBM and ABM are applied to grayscale or color images but we work with high-dimensional images instead. 
Typical ABM involve direct comparison of individual points and thus can be applied to multi-channel data, by directly comparing the multi-dimensional points with any applicable metric.
For FBM, a direct extension of a single-channel feature to multi-channel data does not always exist or a straightforward extension to multiple dimensions suffers from problems.
Consider, for example, local histograms. They work well to summarize single-channel data, but as the number of dimensions of the histogram reflects the number of channels in the data, a local histogram for high-dimensional data would typically be sparse due to the curse of dimensionality~\cite{Altman2018}. %
Therefore, in addition to creating multi-dimensional features, we also consider computing traditional one-dimensional features per channel and then compute the distance between the resulting feature vectors.

In the following, we will present a point cloud distance, namely the Chamfer distance, to showcase an area-based method. 
We use channel-wise histograms as single-channel features and the more general covariance matrix of the neighborhood values as a multi-channel feature to provide examples for FBM.
\autoref{fig:method-overivew} illustrates the concept behind the three approaches.
An example of these methods on synthetic data
will be discussed in more detail in \autoref{sec:synthetic:discussion}.

\subsection{Feature-based methods}\label{sec:method:feature-based}
A wide range of image features exist and an adequate choice depends on the application and the goal of the analysis~\cite{Humeau-Heurtier2019}. 
It is out of the scope of this paper to cover all possibilities.
We focus on spatial heterogeneity which has been successfully applied, for example, in biomedical tumor analysis~\cite{Depeursinge2017} or geo-spatial data analysis~\cite{Fischer2010}.

We investigate two texture features that capture local heterogeneity in scalar images.
As a single-channel feature example, we capture heterogeneity with local histograms per pixel and channel.
Histograms have been successfully used for texture synthesis~\cite{Heeger95} and lend them-self well as texture features.
But since local histograms do not adapt well to high-dimensional data, we use the covariance matrix $\covmat$ and channel-wise means $\meanvec$, 
roughly generalizing the histogram measure of dispersion, as a multi-channel feature. 
Covariance information has as well been shown to be useful texture information for texture synthesis in generative adversarial networks~\cite{gatys2015}.
For FBM, the neighborhood distance becomes
${\spatialDist(\imgPos, \imgPos[j], \imgHighDim, \numSpNeigh)}= {\spatialDist[feat](\featureExtract\ (\imgHighDim, \imgPos ), \featureExtract\ (\imgHighDim, \imgPos[j] ) )}$ with $\featureExtract$\ being the chosen feature extraction operator that will depend on the use case and neighborhood size parameter~$\numSpNeigh$.

It is worth noting that the approach of computing the features separately per channel assumes independence between all channels~$\imgOneDim$, which is typically not the case.
This means that in some cases certain combinations of attribute values and texture features cannot be distinguished.
The covariance matrix feature (\autoref{sec:method:covmat}), and point cloud distance-based, (\autoref{sec:method:area-based}), approaches do not have this limitation since they use the full attribute space to measure the distances.

\subsubsection{Local histograms features} \label{sec:method:vec-histo}
Local histograms are a common way to characterize texture in scalar image processing. 
We compute one feature, i.e., the normalized local histogram, per pixel and channel (confer the right side of \autoref{fig:method-overivew}).
The histogram of attribute values of channel~$c$ in the spatial neighborhood $\spatialNeighborhood_{\imgPos}$ is referred to as the vector~${\histogram_{\imgPos c}=[\histogramBin_{\imgPos c1}, \dots, \histogramBin_{\imgPos cB}]}$, where~$B$ is the total number of bins. 
All entries are normalized by the total amount of pixels in the neighborhood.
As the histogram is represented as a vector, rather than a single scalar, this yields a feature matrix per pixel:
\begin{equation}
    \featureExtract[hist] (\imgHighDim, \imgPos) = %
    [\histogram_{\imgPos 1}, \dots, \histogram_{\imgPos \numAttrDimensions}].
\end{equation}

This means, to compute the distance between two pixels, we now need to compare two vectors of histograms. 
We can interpret a histogram as an estimate of a probability density function.
As such, we can choose one of the many distance functions defined between probability distributions.
One such distance is the \gls*{qf} distance~\cite{Equitz1995}, defined as
\begin{align}
	d^{QF}(\histogram_{\imgPos c}, \histogram_{\imgPos[j] c}) &=
	(\histogram_{\imgPos c} - \histogram_{\imgPos[j] c})^\intercal\ \mathbf{A}\ (\histogram_{\imgPos c} - \histogram_{\imgPos[j] c}) \\
	&= \sum_{b, k = 1}^B a_{bk} (\histogramBin_{\imgPos cb} - \histogramBin_{\imgPos[j] cb}) (\histogramBin_{\imgPos ck} - \histogramBin_{\imgPos[j] ck}).
	\label{eq:QF}
\end{align}
Here,~$\mathbf{A} = \{a_{bk}\}$ with~${0 \leq a_{bk} \leq 1}$ and~${a_{bb} = 1}$, enables attributing a weight between bin indices, e.g., to take distance into account.
We use~${a_{bk} = 1 - (\mid b - k \mid) / B}$ as proposed by Equitz et al.~\cite{Equitz1995}.

With the distance per channel in place, we can define a distance $\spatialDist[feat]$ for all channels as the sum of all channel-wise feature distances:
\vspace{-2mm}\begin{equation}
	\spatialDist[hist](\featureExtract[hist] (\imgHighDim, \imgPos), \featureExtract[hist] (\imgHighDim, \imgPos[j])) = 
	\sum_{c=1}^C d^{QF}(\histogram_{\imgPos c}, \histogram_{\imgPos[j] c}).
	\label{eq:dist:histogram}
\end{equation}\vspace{-4mm}

\begin{figure*}[t]
    \begin{center}
        \vspace{-2mm}
        \includegraphics[width=\linewidth]{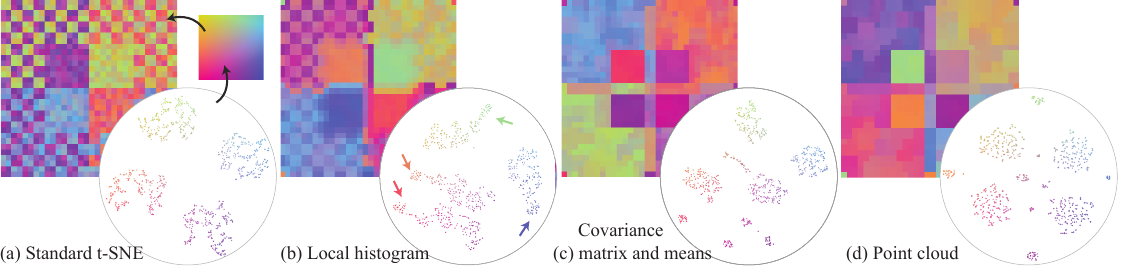}
        \phantomsubcaption\label{fig:ArtExamplesOverview:tsne}  
        \phantomsubcaption\label{fig:ArtExamplesOverview:qf}   
        \phantomsubcaption\label{fig:ArtExamplesOverview:cov}    
        \phantomsubcaption\label{fig:ArtExamplesOverview:pc} 
        \vspace{-7mm}
    \end{center}
    \vspace{-2mm}
    \caption{\textbf{Comparison of different embeddings} of the synthetic image data shown in~\autoref{fig:ArtStructure} using standard t-SNE (a) and the three presented texture-aware approaches (b-d). We use the re-coloring approach introduced in \autoref{fig:ArtStructure} to indicate embedding structure in image space. Here, we use the 2D embedding coordinates to index the colormap shown in (a). As a result, pixels that are close in the embedding space have similar colors in image space.}
    \vspace{-4mm}
    \label{fig:ArtExamplesOverview}
\end{figure*}

\subsubsection{Covariance matrix and means} \label{sec:method:covmat}
A more general dispersion feature for multi-channel data are covariance matrices. 
We will use these in combination with channel-wise mean values as an example for multi-channel features
$\featureExtract[cov] (\imgHighDim, \imgPos) = [\covmat_{\imgPos}, \meanvec_{\imgPos}]$.
Each entry in $\covmat_{\imgPos} = \{ \cov_{a b} \}$ represents the variance between the attribute values within the spatial neighborhood $\spatialNeighborhood_{\imgPos}$ of the channels $\imgOneDim[a]$ and $\imgOneDim[b]$;
the vector $\meanvec_{\imgPos}$ holds the mean values within the same neighborhood per channel. 

One measure that is suited for comparing our covariance matrix feature is the Bhattacharyya distance~\cite{Choi2003}:
\begin{align}
    \spatialDist[Bhat](\featureExtract[cov] (\imgHighDim, \imgPos), \featureExtract[cov] (\imgHighDim, \imgPos[j])) &= 
    \frac{1}{8} (\meanvec_{\imgPos} - \meanvec_{\imgPos[j]})^T \covmat^{-1} (\meanvec_{\imgPos}  - \meanvec_{\imgPos[j]}) \nonumber \\
    &\ + \frac{1}{2} \ln \left( \frac{\det \covmat}{\sqrt{ \det \covmat_{\imgPos}  \det \covmat_{\imgPos[j]}}} \right)
    \label{eq:dist:bat}
\end{align}
with $\covmat = \frac{\covmat_{\imgPos} + \covmat_{\imgPos[j]}}{2}$ and $\det(\covmat)$ denoting the determinant of a matrix $\covmat$.

\subsection{Area-based methods} \label{sec:method:area-based}
A straightforward example ABM is to interpret the attribute vectors of pixels in the spatial neighborhood~$\spatialNeighborhood_{\imgPos}$ as a high-dimensional point cloud, see~\autoref{fig:method-overivew}.
Instead of comparing explicit texture features, we are now computing distances in the high-dimensional space defined by the data attributes directly.
Point cloud distances have been used to compare single-channel images~\cite{Huttenlocher1993} and many naturally extend to higher dimensions since they are based around norms of differences between attribute vectors.
To stay consistent with the previously-established notation, one can think of it as simply defining the feature as the data values in the spatial neighborhood without any transformation, $\featureExtract[PC] (\imgHighDim, \imgPos) = \featureMat_{\imgPos} = \{ \dataPointHighDim_{\imgPos[j]} : \imgPos[j] \in \spatialNeighborhood_{\imgPos} \}$, leading to a point cloud distance ${d^{PC}( \featureMat_{\imgPos}, \featureMat_{\imgPos[j]} )}$.

Multiple distance measures between point clouds exist~\cite{Eichner2013}.
The choice, which one to use, depends on the application that needs to be addressed. 
A commonly used point cloud distance is the Chamfer \mbox{(pseudo-)} distance~\cite{Fan2017}. %
Conceptually, to calculate this point cloud distance $d^{PC}$ between the spatial neighborhoods of two pixels~$\imgPos$ and $\imgPos[j]$, for each point in the spatial neighborhood $\spatialNeighborhood_{\imgPos}$ we find the closest point
in the other neighborhood $\spatialNeighborhood_{\imgPos[j]}$ with respect to a metric (e.g., squared Euclidean distance in our implementation) and average these closest point distances. 
This yields:
\begin{equation}
	\begin{split}
		\spatialDist[PC](\featureMat_{\imgPos}, \featureMat_{\imgPos[j]}) = 
		& \frac{1}{\neighborhodSizeTot} \sum_{\imgPos[q] \in \spatialNeighborhood_{\imgPos}} \min_{\imgPos[p] \in \spatialNeighborhood_{\imgPos[j]}} \lVert \dataPointHighDim_{\imgPos[q]} - \dataPointHighDim_{\imgPos[p]} \rVert^2_2\ + \\
		& \frac{1}{\neighborhodSizeTot} \sum_{\imgPos[p] \in \spatialNeighborhood_{\imgPos[j]}} \min_{\imgPos[q] \in \spatialNeighborhood_{\imgPos}} \lVert \dataPointHighDim_{\imgPos[q]} - \dataPointHighDim_{\imgPos[p]} \rVert^2_2
	\end{split}
	\label{eq:chamfer}
\end{equation}

In comparison with other point cloud distances, like the closely related Hausdorff distance, the Chamfer distance is more robust against outliers in the neighborhoods. Unlike the max-min Hausdorff distance, here, we take the average of all point-wise minima instead of their maximum.
\nameshortref[Supplemental Material ]{supp:pc} compares several Hausdorff family distances for the previously introduced toy-example image discussed in~\autoref{sec:synthetic}.

\subsection{Computational complexity} \label{sec:method:complex}
The computational complexity for the presented approaches can be split into two parts: first the feature extraction and second the actual distance functions.

The histogram feature computation in our implementation scales linearly with the number of spatial neighbors.
For a single pixel and channel, this yields the complexity for the \textbf{local histogram} feature extraction:
$\mathcal{O}(\neighborhodSizeTot)$, scaling linearly, only with the number of pixels in the neighborhood $\neighborhodSizeTot\ = (2 \cdot \numSpNeigh + 1)^2$.
The \textbf{covariance matrix} feature calculation is dominated by the computational complexity of a  matrix multiplication between two matrices of size $\neighborhodSizeTot \times \numAttrDimensions$, namely~$\mathcal{O}(\numAttrDimensions \neighborhodSizeTot^2)$.

The distance calculation is the more time-consuming step for the presented methods.
For the \textbf{local histogram} feature approach the QF distance computation scales quadratically with the number of bins $B$, which dominates its complexity in~$\mathcal{O}(\numAttrDimensions\, B^2)$.
\textbf{Covariance matrices and mean} comparison with the Bhattacharyya distance is more expensive. 
Its computation involves matrix-vector multiplication and determinant calculation.
Using LU decomposition for the latter yields~$\mathcal{O}(\numAttrDimensions^3)$.

Finally, the \textbf{point cloud distance} requires the computation of all pairwise distances between the two neighborhoods.
As a result, it's complexity scales quadratically with the neighborhood size~$\mathcal{O}(\numAttrDimensions\, \neighborhodSizeTot^2)$. 
Note, that neighborhood-based dimensionality reduction methods, like t-SNE and UMAP, use the point distances to construct k-nearest neighbor (knn) graph, which requires the computation of all pairwise distances for the whole dataset.
Naively, the knn graph construction would scale quadratically with the number of pixels rather than the neighborhood. %
However, most modern implementations of t-SNE, and UMAP avoid quadratic complexity by using approximated knn algorithms~\cite{Pezzotti2017AtSNE, mcinnes2018umap}, as do we, see~\autoref{sec:implementation}.

We have also experimentally verified this analysis, showing that the local histogram approach is the fastest and the Bhattacharyya distance the slowest. 
The full data can be found in the~\nameshortref[Supplemental Material ]{supp:timings}.

\subsection{Spatial weighting} \label{sec:method:weights}
So far, we treated all pixels in the spatial neighborhood uniformly.
In order to define specific patterns of interest within the neighborhood, for example by assigning pixels further away from the center a lower importance, we introduce a spatial weighting $\weightsvec$. 
Weights are consistent with respect to the center $\imgPos[i]$ of a neighborhood, which implies that a pixel position $\imgPos[p]$ receives the weight: $\weightsvec(\imgPos[i]-\imgPos[p])$.

For the histogram features, spatial weights with~${\sum \weightsvec = 1}$ can be included in the histogram construction by scaling pixel attributes by the weight.
Weights can be introduced into the covariance matrix and mean computation as detailed in the~\nameshortref[Supplemental Material ]{supp:weights}, where we also cover the integration into the Chamfer point cloud distance. 
A two-dimensional Gaussian kernel is a natural weighting choice as it assigns smoothly decreasing importance to pixels further away from the neighborhood center.

\section{Application on synthetic data} \label{sec:synthetic}

To illustrate some of the properties of the different approaches, presented in~\autoref{sec:method:feature-based} and ~\autoref{sec:method:area-based}, we created a simple synthetic image data set, shown in \autoref{fig:ArtStructure}.
The image consists of $32 \times 32$ pixels, with two attribute channels, separating the pixels into four groups (\autoref{fig:ArtStructure_pc}). 
As seen in \autoref{fig:ArtStructure_image} the spatial layout includes four homogeneous regions in the center and checkered patches around them, each constructed by alternating $2 \times 2$ pixel blocks of two different classes.

\autoref{fig:ArtExamplesOverview} shows a standard t-SNE embedding of the synthetic data set as well as embeddings using the three described methods.
All four embeddings were computed using a perplexity of $20$ and $1,000$ gradient descent iterations. 
For the three texture-aware approaches we considered a uniformly weighted~$3 \times 3 $ neighborhood.
To indicate structure derived from the t-SNE embedding in image-space without clustering, we use a simple re-coloring previously shown by H\"ollt et al.~\cite{ bib:hollt:2019a}.
In short, 2D coordinates derived from the embedding are added to each pixel.
The pixel is then assigned a color by using these coordinates as a lookup into a 2D colormap.
As a result, pixels that are close in the embedding space, and thus are similar according to the used distance metric, will have a similar color in the image representation.
We use t-SNE as an example throughout the paper, similar embeddings using UMAP and MDS can be found in \nameshortref[Supplemental Material~]{supp:UMAPandMDS}. 

\textbf{Standard t-SNE} (\autoref{fig:ArtExamplesOverview:tsne}) separates the pixels into four groups, one per class, with some small scale structure within each class, introduced by noise in the data.
However, the embedding does not give any insight into the spatial layout of the four classes.
In particular, the pixels of each class positioned in the checkerboard pattern cannot be distinguished from pixels of the same class in the central homogeneous regions.

\autoref{fig:ArtExamplesOverview:qf}, shows the embedding and re-coloring using the \textbf{Local histograms and \gls*{qf} distance}.
We defined the number of histogram bins using the Rice rule: $B = \lceil 2 \sqrt[3]{M} \rceil = 5$, with the neighborhood size $M = 3 \times 3 = 9$. 
The resulting embedding is somewhat less clear than the standard t-SNE one, consisting of only three major clusters, however, with more structure within those clusters.
The four homogeneous areas in the center show up in separated areas in the embedding (arrows), indicated by their individual colors.
These regions are loosely connected to larger regions in the embedding containing the pixels from the checkerboard regions.
Notably, the individual two classes forming a checkerboard region do form separate regions within the larger clusters to some degree. 
However, pixels of the same class in two different checkerboard regions, for example, pixels of with small values in both channels~(\autoref{fig:ArtStructure}), which are present in both checkerboards on the left image half, are separated, as indicated by the blue-ish and orange-ish colors.

\begin{figure*}[b!]
    \begin{center}
    \vspace{-1mm}
        \includegraphics[width=\textwidth]{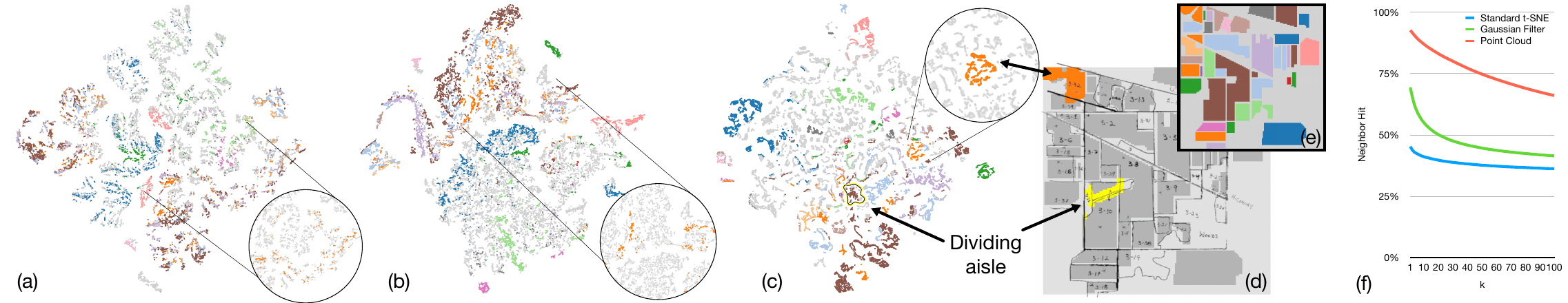}%
        \phantomsubcaption\ignorespaces\label{fig:IndianPines:EmbGtColored:std}
        \phantomsubcaption\ignorespaces\label{fig:IndianPines:EmbGtColored:blur}
        \phantomsubcaption\ignorespaces\label{fig:IndianPines:EmbGtColored:cham}
        \phantomsubcaption\ignorespaces\label{fig:IndianPines:EmbGtColored:aisle}
        \phantomsubcaption\ignorespaces\label{fig:IndianPines:EmbGtColored:gt}
        \phantomsubcaption\ignorespaces\label{fig:IndianPines:EmbGtColored:quant}
        \vspace{-8mm}
    \end{center}
    \vspace{-1mm}
    \caption{\textbf{Indian Pines: Comparison of texture-aware and standard t-SNE embeddings.}
    Embeddings using standard t-SNE (a), standard t-SNE applied after Gaussian filtering (b), and our point cloud based t-SNE (c).
    A manually annotated map and ground truth labels are shown in (d) and (e), respectively.
    Points in the embeddings are colored, according to ground truth labels for a qualitative comparison of embedding structure.
    Finally, we show the ration of k-nearest neighbors in embedding space with the same ground truth label for different k in (f).
    Note, that for an exploratory use-case no ground truth is available and we only show it here to illustrate the properties of the embeddings.
    \vspace{-6mm}
    }
    \label{fig:IndianPines:EmbGtColored}
\end{figure*}

The result of our approach using the \textbf{covariance matrix and means feature} can be seen in \autoref{fig:ArtExamplesOverview:cov}. 
We can clearly identify nine separate clusters in the embedding.
The four homogeneous regions correspond to the four small clusters on the bottom of the embedding, while the four larger clusters represent the four checkerboard regions, as indicated by the recolored image.
The ninth cluster corresponds to the boundaries between the checkered regions.
Again, the homogeneous areas are separated from the checkered but with much sharper boundaries.
Different from the previously described approaches, however, each checkered area is recognized as a single cluster in the embedding, meaning that the checkerboard pattern is not visible anymore in the re-colored image. 
In an exploratory visual analysis setting, this would facilitate the selection and further analysis of regions with specific spatial neighbourhood characteristics.

The \textbf{Point cloud distance} approach, here specifically using the Chamfer distance, yields a similarly straightforward partitioned embedding  in~\autoref{fig:ArtExamplesOverview:pc}.
Again, all four homogeneous image patches are separately clustered as well as the checkered regions.
The borders between the different regions now also created individual clusters in the middle of the embedding.

To quantitatively analyse the approaches, we compute the $k$-nearest neighbor hit, as described by Espadoto et al~\cite{Espadoto2019}.
The average neighbor hits for 63-nearest neighbors in embedding space are $77.9\%$ (point cloud distance), $79.4\%$ (Bhattacharyya distance) and $80.4\%$ (QF distance) whereas the standard t-SNE embedding yields $35.1\%$.
While the point cloud and Bhattacharyya distance result in a higher neighbor hit for small neighbor numbers, their quality decreases slightly faster for larger numbers than the QF distances hit.
See~\nameshortref[Supplemental Material ]{supp:settings} and \autoref{fig:supp:UMAPandMDS:quant:tsne} in ~\nameshortref[Supplemental Material ]{supp:UMAPandMDS} for full details.

\subsection{Discussion} %
\label{sec:synthetic:discussion}

All three presented texture-aware approaches are able to distinguish between several spatial arrangements of the high-dimensional image.

A drawback of local histogram features is the number of bins as an additional hyperparameter.
Since there is no obvious choice for a good setting, the user has to fall back to heuristically setting this value.
Further, in the scope of this work, we only discuss the \gls*{qf} distance for comparing histograms.
Other distance measures are available and would likely produce different results.
As a per-channel feature-based approach the local histograms implicitly assume channel independence. 
Thus, they cannot capture multidimensional texture patterns. 
Using multidimensional histograms instead of a 1D histogram per channel might be able to capture such patterns.
However, such an approach would drastically increase computational complexity.
The histogram size grows exponentially with the number of dimensions, quickly making storing histograms and computing distances infeasible. 
Further, such histograms are in danger of quickly becoming very sparse and as such would not provide a useful basis for comparison anymore.

The covariance matrix feature can capture multiple attribute dimensions without requiring channel independence with the same goal of comparing the distribution of values within the defined neighborhood. 
However, instead of comparing all individual values, they are represented in an approximate way based on the assumption that the values are Gaussian distributed. 
The point cloud distance does not make this assumption and compares all attribute vectors to each other.
If we compare the covariance matrix feature with the point cloud distance results, the most prominent difference between the embeddings is how they treat the borders between cluster regions.
When data has a bi-modal distribution in a channel, the Gaussian assumption in the covariance feature does not reveal this case, whereas the point cloud distance would.

An advantage of the FBM methods is that they produce features that can aid the interpretation of structure in the resulting embeddings.
Visualizing those features in combination with the embeddings is an interesting avenue for future work.

The spatial weights as introduced in~\autoref{sec:method:weights} and spatial neighborhood size~$\numSpNeigh$ affect the approaches to different degrees.
See the~\nameshortref[Supplemental Material ]{supp:weighting} for a brief overview of different neighborhood sizes and spatial weights.
For example, the Chamfer point cloud distance produces very similarly clustered embeddings for several neighborhood sizes and is \textemdash\ with respect to the synthetic data set \textemdash\ not much affected by radially decreasing spatial weights for this example.
Meanwhile, using the histogram feature, the checkered pixels are clustered differently when weights are applied.

\section{Implementation}  \label{sec:implementation}

We implemented the described distance measures as an extension for the open-source t-SNE implementation in HDI~\cite{Pezzotti2017AtSNE, bib:2019_vis_GPUtSNE},
where we use HNSW~\cite{Malkov2020} with our custom distance functions to create the approximated k-nearest neighbor graph.
The framework is implemented in C++ and OpenGL for GPU-based calculations; a Python wrapper is provided as well.
Our library is available as open-source on GitHub~\cite{bib:githubalex}.

\section{Use cases} \label{sec:useCases}
Here, we illustrate the application of the presented approaches for visual data exploration using two use cases.
The first use case (\autoref{sec:case:hyperspectral}) describes the exploration of hyperspectral images, commonly used in geospatial analysis.
For the second use case (\autoref{sec:case:imc}), we applied our method to imaging mass cytometry data, a method that is recently gaining attention in systems biology.
We use the Chamfer point cloud distance as an example for an area-based method and the covariance matrix feature exemplarily as a multi-channel feature-based method.
The Bhattacharyya distance comes with a high computational cost for high large channel numbers, hence the usage of the point cloud distance for the hyperspectral image example and the covariance matrix feature for the systems biology use case.

\subsection{Hyperspectral imaging}\label{sec:case:hyperspectral}
Similar to digital photography, hyperspectral imaging captures bands of the electromagnetic spectrum.
Instead of only three channels for red, green, and blue in digital photography, hyperspectral imaging captures hundreds of narrow spectral bands at the same time, each corresponding to a channel in the output image. 
This spectral signature can be used to recognize materials or objects in the image.
To fully exploit the information in hyperspectral images, spatial relations between the high-dimensional pixels need to be considered~\cite{Ghamisi2017}.

\begin{figure*}[!t]
    \begin{center}
        \vspace{-4mm}
        \includegraphics[width=\textwidth]{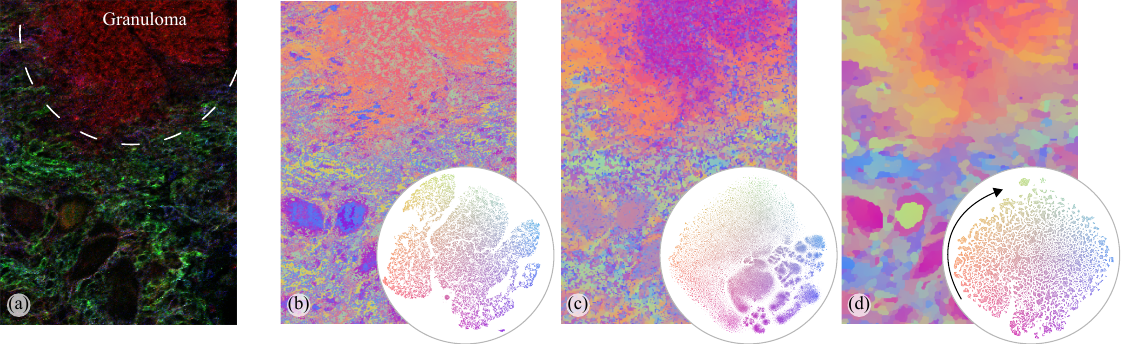}%
        \phantomsubcaption\ignorespaces\label{fig:SingleCell:GranulomeOverview:color}
        \phantomsubcaption\ignorespaces\label{fig:SingleCell:GranulomeOverview:std}
        \phantomsubcaption\ignorespaces\label{fig:SingleCell:GranulomeOverview:bat_k1}
        \phantomsubcaption\ignorespaces\label{fig:SingleCell:GranulomeOverview:bat_k3}
        \vspace{-10mm}
    \end{center}
    \caption{\textbf{Imaging mass cytometry.}
    A false coloring image of a lung tissue sample gives an idea of the tissue structure in (a) with the granuloma enclosed by the dashed line.
    Embeddings derived using standard t-SNE (b), and texture-informed embeddings using the covariance matrix and Gaussian weighting with a $3 \times 3$ (c) and $7 \times 7$-sized (d) neighborhood are shown in combination with recolored images to indicate embedding structure in image space.
    \vspace{-6mm}
    }
    \label{fig:SingleCell:GranulomeOverview}
\end{figure*}

Here, we present a use case, based on the Indian Pines data set~\cite{baumgardner2015220}.
The data set is an established reference hyperspectral image obtained by airborne visible/infrared imaging spectrometry, covering 220 adjacent spectral bands, of which 200 are typically used (after discarding 20 water absorption bands that do not contain useful information).
We consider a $145 \times 145$ pixel cutout of the data set, known as \emph{Site~3}.
This subset of the data is one of three \emph{`intensive'} test sites and thus has been well documented.
A ground truth (\autoref{fig:IndianPines:EmbGtColored:gt}), providing labels indicating ground usage, such as fields, grassland, or houses for all pixels is available for the data set.
We use this ground truth data to verify that the structure in our embeddings is meaningful; however, it must be noted that in explorative analysis such a ground truth is not available and information would need to be derived from the embedding structure.
Each pixel in the data set maps to a $20 \times 20$ meter area on the ground.
The 200 channels of this cutout cover wavelengths range from to \SIrange{400}{1300}{\nano\meter} in roughly \SIrange{9}{10}{\nano\meter} steps.

We computed embeddings using standard t-SNE (\autoref{fig:IndianPines:EmbGtColored:std}) and our texture-aware t-SNE using the Chamfer point cloud distance with a $5 \times 5$ neighborhood (\autoref{fig:IndianPines:EmbGtColored:cham}).
To provide a better baseline than standard t-SNE, we applied a Gaussian filter to each channel of the original image data and derived a standard t-SNE embedding from the filtered image (\autoref{fig:IndianPines:EmbGtColored:blur}).
A Gaussian filter applies smoothing to an image by convolving it with a Gaussian kernel.
Thus in the resulting image every pixel is a combination of a small neighborhood, providing a  straightforward way of incorporating some spatial neighborhood information.
We computed all embeddings using a perplexity of 30 and $5,000$ gradient descent iterations.
In \autoref{fig:IndianPines:EmbGtColored}, we color-code the ground truth labels on each embedding to indicate how well the structure in the embedding corresponds to structure in the images.

The embedding based on the Chamfer point cloud distance shows more structure than the other two embeddings. 
Notably, the colored points, corresponding to the labels of the ground truth, form more clearly distinguished clusters (see, for example, the orange points in the insets of~\autoref{fig:IndianPines:EmbGtColored}).
The other two embeddings show many clusters containing points belonging to multiple regions.
Most notably is the weak separation of the background, unlabeled points (light gray) in \Autoref{fig:IndianPines:EmbGtColored:std, fig:IndianPines:EmbGtColored:blur}, compared to the coherent, strongly separated groups in the point cloud-based embedding (\autoref{fig:IndianPines:EmbGtColored:cham}).

This visual impression is reinforced by a quantitative analysis using the neighborhood hit~\cite{Espadoto2019}, the average ratio of $k$-nearest neighbors in embedding space with the same ground truth label.
\autoref{fig:IndianPines:EmbGtColored:quant} shows the neighborhood hit for the first $100$ nearest neighbors.
The point cloud distance approach yields a significantly higher hit for all $k$-nearest neighbors than both standard t-SNE and the Gaussian filtering approach.

\begin{figure*}[h]
    \centering
    \begin{minipage}{0.45\textwidth}
        \begin{subfigure}{0.45\textwidth}
             \centering
             \includegraphics[width=\textwidth]{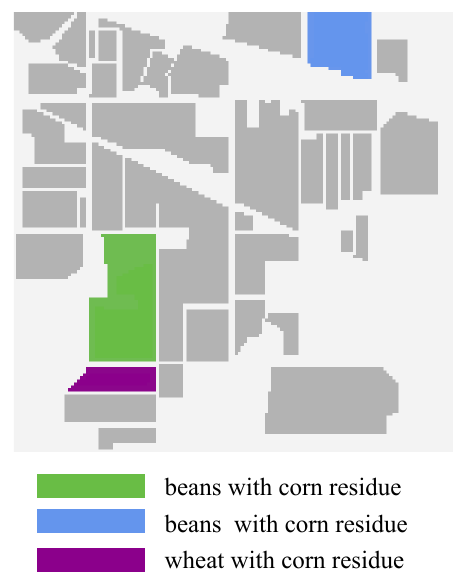}
             \caption{Highlights in the ground truth.}
             \label{fig:IndianPinesGroundTruth}
         \end{subfigure}
         \hfill
        \begin{subfigure}{0.5\textwidth}
             \centering
             \includegraphics[width=\textwidth]{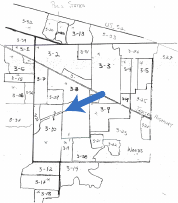}
             \caption{Dividing aisle.}
             \label{fig:IndianPines:aisle}
         \end{subfigure}
    \caption{Ground truth of Indian Pines Site 3 with highlighted areas as further discussed in~\autoref{fig:IndianPines:EmbSelCompFacet} and an annotated map of the Site without a color-coded background.}
    \label{fig:IndianPines:gtAndaisle}
    \end{minipage}\hfill
    \begin{minipage}{0.54\textwidth}
        \centering
        \includegraphics[width=1\textwidth]{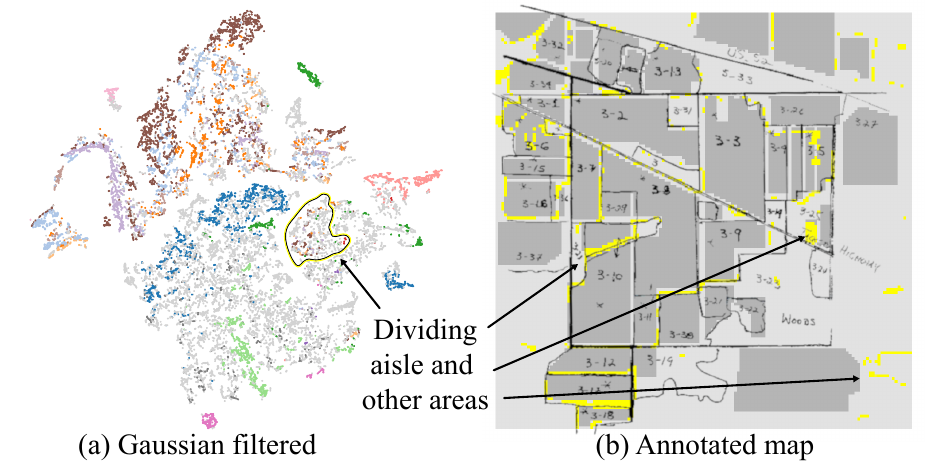} %
        \caption{Attempted selection of the dividing aisle in the t-SNE embedding based on the Gaussian filtered data. In contrast to~\autoref{fig:IndianPines:EmbGtColored:aisle}, it is not possible to find a cluster that only corresponds to the aisle.}
        \label{fig:IndianPines:gtAndaisleBlurred}
    \end{minipage}
\end{figure*}

\autoref{fig:IndianPines:EmbGtColored:aisle} shows the original hand-drawn annotations overlaid on the individual fields taken from the ground-truth data.
The arrow points at an aisle dividing two parts of a field that was given in the manual annotation but was lost in the ground truth.
In our point cloud-based embedding, we could identify a cluster (arrow in \autoref{fig:IndianPines:EmbGtColored:cham}) corresponding to this aisle and an unlabeled area next to it.
The yellow area in \autoref{fig:IndianPines:EmbGtColored:aisle} indicates the pixels corresponding to that cluster, which illustrates the ability to distinguish structure, even beyond the ground truth, in the case of the embedding using the point cloud distance. 
Hereby, we illustrate the usefulness of combining spatial information with the full attribute space for exploration purposes.
While there exist clusters in the other two embeddings that partially correspond to the aisle, they also contain pixels from areas in different regions (see \autoref{fig:IndianPines:gtAndaisleBlurred} in~\nameshortref[Supplemental Material ]{supp:pines} for an example).

\begin{figure*}[h]
    \begin{center}
        \includegraphics[width=\textwidth]{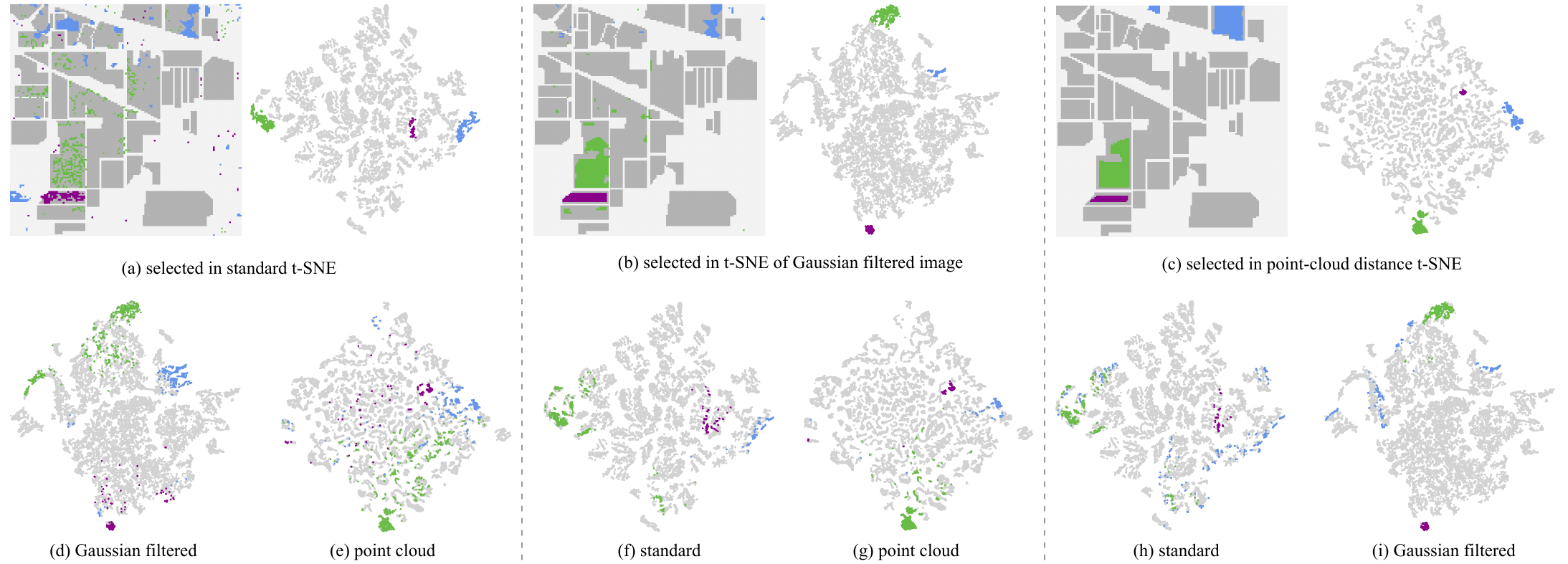}%
        \phantomsubcaption\ignorespaces\label{fig:IndianPines:EmbSelCompFacet:tsne}
        \phantomsubcaption\ignorespaces\label{fig:IndianPines:EmbSelCompFacet:blur}
        \phantomsubcaption\ignorespaces\label{fig:IndianPines:EmbSelCompFacet:cham}
        \phantomsubcaption\ignorespaces\label{fig:IndianPines:EmbSelCompFacet:blurINtsne}
        \phantomsubcaption\ignorespaces\label{fig:IndianPines:EmbSelCompFacet:chamINtsne}
        \phantomsubcaption\ignorespaces\label{fig:IndianPines:EmbSelCompFacet:tsneINblur}
        \phantomsubcaption\ignorespaces\label{fig:IndianPines:EmbSelCompFacet:chamINblur}
        \phantomsubcaption\ignorespaces\label{fig:IndianPines:EmbSelCompFacet:tsneINcham}
        \phantomsubcaption\ignorespaces\label{fig:IndianPines:EmbSelCompFacet:blurINcham}
        \vspace{-8mm}
    \end{center}
    \caption{\textbf{Indian Pines: Comparison of spatially-aware and standard t-SNE embeddings.} 
    The top row (a-c) shows a standard t-SNE embedding, an embedding of the Gaussian filtered data set and our spatially-aware embedding (based on the Chamfer point cloud distance). We tried to select the three highlighted regions from~\autoref{fig:IndianPinesGroundTruth} in each embedding and show the respective pixel on the ground truth. The lower row (d-i) highlights the selections made in the above embeddings in the two other embeddings, for example highlighted points in (d) and (e) correspond to points selected in (a).}
    \label{fig:IndianPines:EmbSelCompFacet}
\end{figure*}

In summary, our point-cloud embedding outperforms the other two with respect to the exploration of spatially continuous, meaningful regions. 
More examples for similar behaviour, for instance that a specific field is well captured in a single cluster of the point-cloud embedding but divided between multiple clusters in the other embeddings, are shown in~\Autoref{fig:IndianPines:EmbSelCompFacet} in~\nameshortref[Supplemental Material ]{supp:pines}.

\subsection{Imaging mass cytometry}\label{sec:case:imc}

Imaging mass cytometry~\cite{Giesen2014} is a recent imaging modality
used to study cellular biology.
Imaging mass cytometry simultaneously captures the expression of up to 50 different proteins in tissue by ablating tissue sections spot-by-spot.
Combining the resulting measurements in a regular grid results in a high-dimensional image, where the pixel position corresponds to the position in the tissue and the channels to the different measured proteins.
Visual analytics and exploratory analysis based on dimensionality reduction is used in practice for the analysis of imaging mass cytometry data, as for example presented by Somarakis et al.~\cite{Somarakis2019}.
They present a multi-step approach where cells are segmented, followed by aggregating high-dimensional profiles per cell.
These are then used to identify cell types using dimensionality reduction and the resulting classification is the basis for exploration of local neighborhoods of cells.

For this use case, we consider a tissue sample from a mammalian lung provided by collaborating researchers.
The image measures $272 \times 374$ pixels, each pixel represents a~\SI{1}{\micro\meter}~area, and we consider ten attribute channels, corresponding to
ten different proteins describing immune cells and structural properties. 
\autoref{fig:SingleCell:GranulomeOverview} shows an RGB re-coloring of the sample, mapping red, green, and blue channels to one of three structural proteins each, to give an impression of the tissue layout.

\autoref{fig:SingleCell:GranulomeOverview:std} shows a standard t-SNE embedding and re-colored following the same re-coloring scheme as in \autoref{fig:ArtExamplesOverview}.
\Autoref{fig:SingleCell:GranulomeOverview:bat_k1, fig:SingleCell:GranulomeOverview:bat_k3} show texture-aware embeddings based on the covariance matrix feature and Bhattacharyya distance.
We used two different neighborhood sizes, $3 \times 3$ in \autoref{fig:SingleCell:GranulomeOverview:bat_k1} and $7 \times 7$ in \autoref{fig:SingleCell:GranulomeOverview:bat_k3} to show structures on different scales in the image.
For both, we make use of a Gaussian spatial weighting.
All embeddings were computed with a perplexity of 30 and $5,000$ gradient descent iterations to ensure convergence.

Our collaborators are interested in the composition of cell structures called granuloma, indicated in \autoref{fig:SingleCell:GranulomeOverview}, and their surrounding cells.
A granuloma is an agglomeration of immune cells, typically to isolate irritants or foreign objects.
Current analysis pipelines separate the analysis of the high-dimensional attribute data and spatial layout of the cell data~\cite{Somarakis2019}. 
Our collaborators stated that it would be useful to combine these two steps for early data exploration.

In the re-colored image in \autoref{fig:SingleCell:GranulomeOverview:std} we can see that the granuloma as a whole is already differentiated from the rest of the image, indicated by a bright orange area with mint green, purple, and blue inclusions.
The bright orange indicates a combination of proteins, characteristic for a specific set of immune cells (macrophages) which are expected in the center of the granuloma. %
It is known that the area around a granuloma is made up of layers comprising of said macrophages
and different combinations of other immune cells.
The structure of these cell layers and interaction of cells within and between adjacent layers is subject of current research.
Hints of this changing composition can be seen in \autoref{fig:SingleCell:GranulomeOverview:std} where the center largely consists of mint-green inclusions which slowly change to purple and blue inclusions towards the outside.
A first hypothesis when analysing the given tissue was that these layers are similar all around the granuloma.

Comparing the small-scale texture-aware embedding and re-colored image in \autoref{fig:SingleCell:GranulomeOverview:bat_k1}, we get a similar impression with a bright-purple colored area with some inclusions defining the granuloma. 
Note, that the colors are not directly comparable due to the heuristic nature of t-SNE and the different structures of the embeddings.
However, we already see some hints at larger scale structure.
The central area of the granuloma (consisting of many blue (mint in \autoref{fig:SingleCell:GranulomeOverview:std}) inclusions) and the outer layers are now separately colored in a deep pink and orange, respectively, indicating separation in the embedding.
Individual cells can still roughly be identified, for example the blue and purple patches within and around the structure.

Finally, using a larger neighborhood as in \autoref{fig:SingleCell:GranulomeOverview:bat_k3} clearly creates areas of similar color, corresponding to higher-level structures. 
This is expected as the neighbourhood is enlarged. %
Here, no individual cells are recognizable anymore.
The granuloma as a whole is clearly recognizable by a pink to orange area, but in addition a clear layering structure is visible.
The granuloma center is a relatively homogeneous dark pink area. %
Around the granuloma, we can see the layering of structures in different shades of orange to a greenish tone on the far outside, following the colormap applied to the embedding from bottom to top on the left side (arrow \autoref{fig:SingleCell:GranulomeOverview:bat_k3}).
Upon inspection of this texture-aware embedding our collaborators were very interested in these layers surrounding the granuloma center and how clearly they were identifiable in \autoref{fig:SingleCell:GranulomeOverview:bat_k3}, hereby eliminating the need for a multi-step approach which was typical for their work flow.
They also noted that the layer structure was more varied than they expected which they intend to study further and verify that this is indeed consistent across biological replicates.

\section{Conclusion}\label{sec:conclusion}

We have presented a framework of texture-aware dimensionality reduction for visual exploration of high-dimensional images and illustrated its potential through examples based on t-SNE and three different texture-aware distance metrics.
The generated embeddings combine attribute similarity with spatial context, and, thereby, support the exploration of high-dimensional images.
Our method adapts the point similarity calculation of distance-based dimensionality reduction methods by taking the spatial nature of images into account. 
We presented two classes of approaches for comparing spatial pixel neighborhoods and extended them to high-dimensional images:
Feature-based methods (FBM), extracting and comparing features of neighborhoods, and area-based methods (ABM), applying distance measures between the sets of attributes within the neighborhoods directly.
We have shown strengths and weaknesses of the different approaches, illustrated them in a synthetic example and presented their applicability via two use cases.

The presented method opens several avenues for future work.
We focused on images, i.e., structured, rectangular grids.
Extensions to unstructured grids, common in geographic information systems, or graphs are thinkable.
While we show several examples of different neighborhood sizes, we did not investigate this parameter in-depth.  
Using varying neighborhood sizes might reveal spatial structures that are only present at a specific scale. 
Another interesting avenue might be to investigate the potential of other feature extraction methods, like Markov random field texture models or neural network approaches to capture domain-specific texture characteristics when training data is available.
Additionally, visualizing the extracted features alongside the embeddings is an interesting idea in itself.

In this work, we have shown that texture-aware dimensionality reduction methods can provide insights into high-dimensional images that cannot be captured with standard dimensionality reduction methods alone. 
Thus, we believe that they will prove to be a valuable addition to the tool box for high-dimensional data analysis.

\acknowledgments{
This work received funding from the NWO TTW project 3DOMICS (NWO: 17126).
We thank Paula Niewold and Simone Joosten (Leiden University Medical Center, LUMC) for providing imaging mass cytometry data and feedback. We also thank Nannan Guo, Frits Koning (LUMC) and Na Li (Jilin University) for valuable discussions.
}

\bibliographystyle{abbrv-doi-hyperref-narrow}
\bibliography{references}

\clearpage
\setcounter{page}{1}
\onecolumn
\begin{center}
{
\makeatletter
\sffamily\ifvgtcjournal\huge\else\LARGE\bfseries\fi\vgtc@sectionfont
Supplemental Material:\\\vspace{5pt}
\vgtc@title
\makeatother
\vspace{14pt}
}

A. Vieth, A. Vilanova, B. Lelieveldt, E. Eisemann, and T. Höllt
\vspace{14pt}

\end{center}

\section*{S0: Computation settings} \labelshort[S0]{supp:settings} %
For all t-SNE computations we set used the following HDLib parameters:
\texttt{Exaggeration=250}, \texttt{exponential~decay=40}, \texttt{number~of~trees=4}, and \texttt{number~of~checks=1024}.
When computing approximated nearest neighbors with HNSWlib we use the default parameters~\texttt{M=16} and \texttt{ef\_construction=200} as well as the random seed 0.

We apply Gaussian filtering using OpenCV~\citesupplement{supp:opencv_library} with the function \texttt{GaussianBlur()} and the settings \texttt{sigmaX=5} and \texttt{ksize=3}.

For the quantitative analysis, we compute the $k$-nearest neighbor hit, as described by Espadoto et al~\citesupplement{supp:Espadoto2019}. 
In brief, for labelled data, for every point in the low-dimensional embedding, we compute the fraction of the $k$ nearest neighbors in the low-dimensional embedding have the same label as the probed point.
This fraction is then averaged for all points in the dataset. 
For the synthetic data, we define the ground truth by separating the checkered areas and homogeneous areas as shown in \autoref{fig:HausdorffFamilyQuant:groundtruth}. 
For the Indian Pines dataset, we use the 16-class ground truth data provided with the original data.
For the synthetic data, we limit $k$ to $k=[1..63]$, as the inner, homogeneous squares in the image cover 64 pixels, meaning larger values for $k$ would include more neighbors than pixels existing for the given label.
For the Indian Pines data, we compute the $k$-nearest neighbor hit for $k=[1..100]$.
For all, we do not include the probed point in the $k$-nearest neighbors.

\section*{S1: Weighted feature computation and weighted Chamfer distance} \labelshort[S1]{supp:weights} %

With weights~$\weightsvec$ that sum to 1 and weighted $\meanvec^{*} = [\mean^{*}_1, \dots, \mean^{*}_C]$, an entry $\cov_{jk}$ of the covariance matric $\covmat_{\imgPos}$ is given by:
\begin{equation}
    \cov_{jk} = \sum_{\imgPos[q] \in \spatialNeighborhood_{\imgPos}} \weightsvec(\imgPos[i]-\imgPos[q]) (\dataPointScalar_{\imgPos[q] j} - \mean^{*}_{j}) (\dataPointScalar_{\imgPos[q] k} - \mean^{*}_{k})^T .
    \label{eq:supp:BC_weights}
\end{equation}
where the weighted means are $\mean^{*}_c = \sum_{\imgPos[q] \in \spatialNeighborhood_{\imgPos}} \weightsvec(\imgPos[i]-\imgPos[q]) \dataPointScalar_{\imgPos[q] c}$.

For weighting the covariance matrix feature, including the channel-wise means, one only needs to introduce the weights~$\weightsvec$ in the calculation of the expected value as probabilities.

The Chamfer point cloud distance from~\autoref{eq:chamfer} can be extended by weighting the minimal distances from each point in the first to the second neighborhood as shown in Equation:
\begin{equation}
	\begin{split}
		\spatialDist[PC](\featureMat_{\imgPos}, \featureMat_{\imgPos[j]}) = 
		& \frac{1}{| \spatialNeighborhood_{\imgPos} |} \sum_{\imgPos[q] \in \spatialNeighborhood_{\imgPos}} \weight(\imgPos[i]-\imgPos[q]) \min_{\imgPos[p] \in \spatialNeighborhood_{\imgPos[j]}} \lVert \dataPointHighDim_{\imgPos[q]} - \dataPointHighDim_{\imgPos[p]} \rVert^2_2\ + \\
		&  \frac{1}{| \spatialNeighborhood_{\imgPos[j]} |} \sum_{\imgPos[p] \in \spatialNeighborhood_{\imgPos[j]}} \weight(\imgPos[j]-\imgPos[p]) \min_{\imgPos[q] \in \spatialNeighborhood_{\imgPos}} \lVert \dataPointHighDim_{\imgPos[q]} - \dataPointHighDim_{\imgPos[p]} \rVert^2_2 .
	\end{split}
	\label{eq:supp:chamfer_weights}
\end{equation}

\section*{S2: Texture-aware UMAP and MDS embeddings} \labelshort[S2]{supp:UMAPandMDS} %
It is possible to use the spatially informed distances between image patches of high-dimensional images in any distance-based dimensionality reduction method.
Here, we show spatially informed UMAP and metric MDS embeddings for the synthetic data set from~\autoref{sec:synthetic}.

We use the umap-learn~\citesupplement{supp:mcinnes2018umap-software} implementation for UMAP and scikit-learn~\citesupplement{supp:scikit-learn} metric MDS.
Note, that in the MDS Bhattacharyya example, the central cluster is actually two: 
the upper part corresponds to the upper left area in the image and the lower part to the lower right. Between the two clusters are the border points between the checkered regions (and the pixels on the vertical border between the homogeneous areas).

\begin{figure*}[ht]
    \centering
    \includegraphics[width=\textwidth]{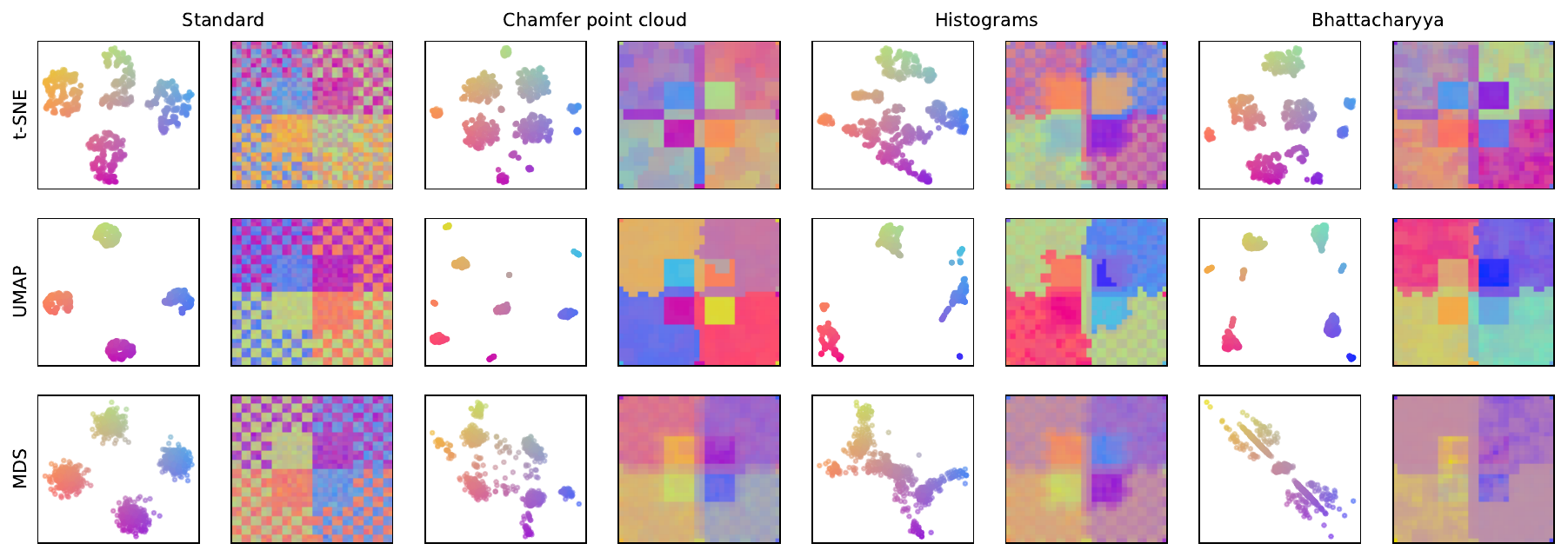}

    \caption{Spatially informed and standard embeddings with t-SNE, UMAP and MDS. Embeddings and recolored images as described in~\autoref{fig:ArtExamplesOverview} in the main paper.}
    \label{fig:supp:UMAPandMDS}
\end{figure*}

\begin{figure*}[!h]
    \centering

    \begin{subfigure}[b]{0.15\textwidth}
        \includegraphics[width=\textwidth]{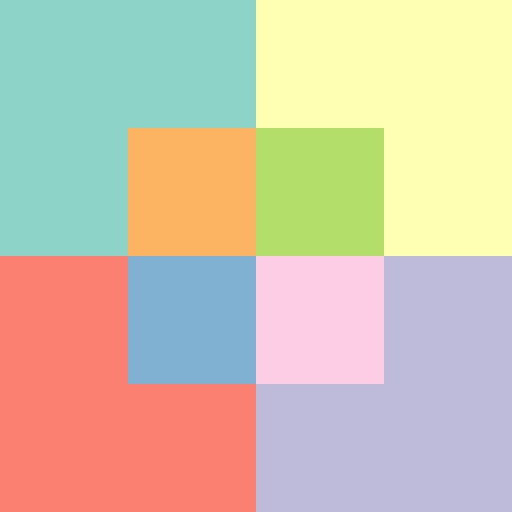}
        \caption{Ground truth.}
        \label{fig:HausdorffFamilyQuant:groundtruth}        
    \end{subfigure}
    \hfill
    \begin{subfigure}[b]{0.275\textwidth}
        \includegraphics[width=\textwidth]{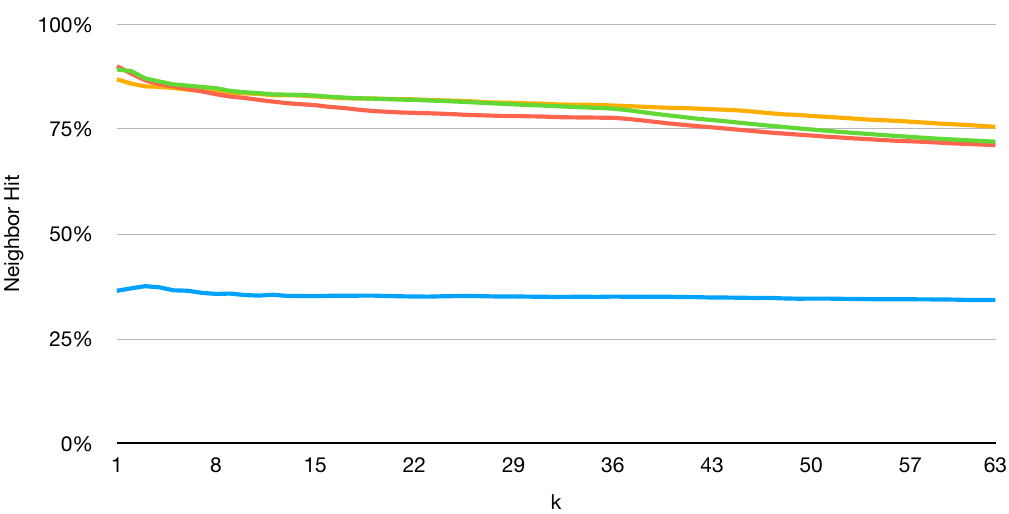}
        \caption{t-SNE}
        \label{fig:supp:UMAPandMDS:quant:tsne}       
    \end{subfigure}
    \hfill
    \begin{subfigure}[b]{0.275\textwidth}
        \includegraphics[width=\textwidth]{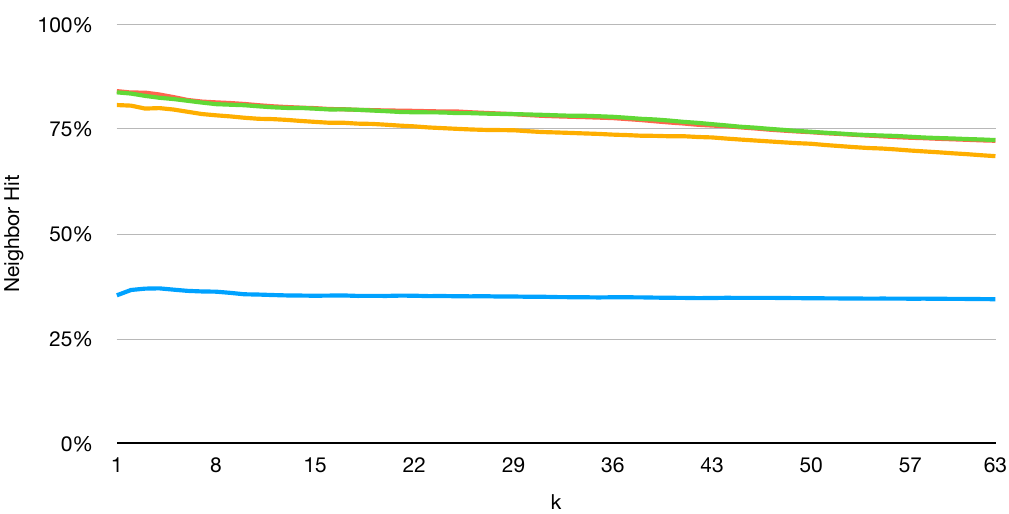}
        \caption{UMAP}
        \label{fig:supp:UMAPandMDS:quant:umap}     
    \end{subfigure}
    \hfill
    \begin{subfigure}[b]{0.275\textwidth}
        \includegraphics[width=\textwidth]{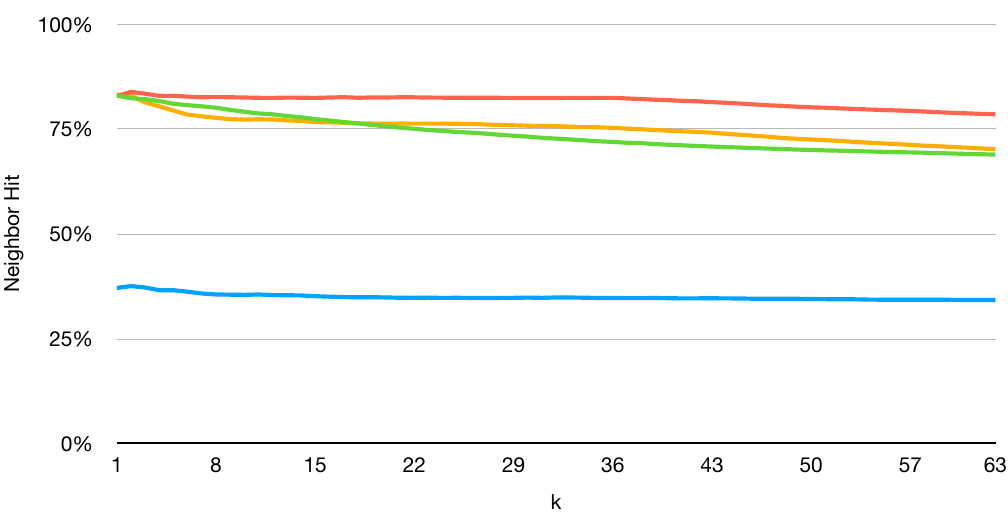}
        \caption{MDS}
        \label{fig:supp:UMAPandMDS:quant:mds}      
    \end{subfigure}
    
    \caption{Ground truth for the synthetic dataset, classifying the four checkered areas on the outside and the four homogeneous regions in the center (a). Nearest neighbor hit for the t-SNE (b), UMAP (c), and MDS (d) embeddings from \autoref{fig:supp:UMAPandMDS}.  \textcolor{plotcolorblue}{\textbf{\textemdash}} standard version, \textcolor{plotcolorgreen}{\textbf{\textemdash}} covariance matrix and mean, \textcolor{plotcolorred}{\textbf{\textemdash}} point cloud distance, \textcolor{plotcoloryellow}{\textbf{\textemdash}} histogram.}
        \label{fig:supp:UMAPandMDS:quant} 
\end{figure*}

\FloatBarrier
\section*{S3: Other point cloud distances} \labelshort[S3]{supp:pc} %

The Chamfer point cloud distance, see \autoref{eq:chamfer}, belongs to the broader family of distances related to the Hausdorff distance. 
These distances build on finding the nearest point for each point in one set to the other. 
Instead of averaging the minima, the Hausdorff distance takes their maximum instead:
\begin{equation}
	d^{PC}_{Haus}(\featureMat_{\imgPos}, \featureMat_{\imgPos[j]}) = \max \biggl\{ 
	\max_{\imgPos[q] \in \spatialNeighborhood_{\imgPos}} \left( \min_{\imgPos[p] \in \spatialNeighborhood_{\imgPos[j]}} \lVert \dataPointHighDim_{\imgPos[q]} - \dataPointHighDim_{\imgPos[p]} \rVert^2_2\ \right),
	\max_{\imgPos[q] \in \spatialNeighborhood_{\imgPos[j]}} \left( \min_{\imgPos[p] \in \spatialNeighborhood_{\imgPos}} \lVert \dataPointHighDim_{\imgPos[q]} - \dataPointHighDim_{\imgPos[p]} \rVert^2_2 \right)  \biggl. \biggr\}
	\label{eq:sup:haus}
\end{equation}

Another variant that might be more robust against outliers in the data might take the median, instead of the average, like the Chamfer distance:
\begin{equation}
	d^{PC}_{HM}(\featureMat_{\imgPos}, \featureMat_{\imgPos[j]}) = \frac{1}{2} \left(
	\operatorname*{median}_{\imgPos[q] \in \spatialNeighborhood_{\imgPos}} \left( \min_{\imgPos[p] \in \spatialNeighborhood_{\imgPos[j]}} \lVert \dataPointHighDim_{\imgPos[q]} - \dataPointHighDim_{\imgPos[p]} \rVert^2_2\ \right) +\ 
	\operatorname*{median}_{\imgPos[q] \in \spatialNeighborhood_{\imgPos[j]}} \left( \min_{\imgPos[p] \in \spatialNeighborhood_{\imgPos}} \lVert \dataPointHighDim_{\imgPos[q]} - \dataPointHighDim_{\imgPos[p]} \rVert^2_2 \right) \right) 
	\label{eq:sup:haus_med}
\end{equation}

Sum of squared differences: When taking the average instead of the minimum of point-wise distances in the Chamfer distance, ending up an average of averages.

\begin{equation}
	\begin{split}
		d^{PC}_{SSD}(\featureMat_{\imgPos}, \featureMat_{\imgPos[j]})
		& = \frac{1}{| \spatialNeighborhood_{\imgPos} |} \sum_{\imgPos[q] \in \spatialNeighborhood_{\imgPos}} \frac{1}{| \spatialNeighborhood_{\imgPos[j]} |} \sum_{\imgPos[p] \in \spatialNeighborhood_{\imgPos[j]}} \lVert \dataPointHighDim_{\imgPos[q]} - \dataPointHighDim_{\imgPos[p]} \rVert^2_2\ + 
		\frac{1}{| \spatialNeighborhood_{\imgPos[j]} |} \sum_{\imgPos[q] \in \spatialNeighborhood_{\imgPos[j]}} \frac{1}{| \spatialNeighborhood_{\imgPos} |} \sum_{\imgPos[p] \in \spatialNeighborhood_{\imgPos}} \lVert \dataPointHighDim_{\imgPos[q]} - \dataPointHighDim_{\imgPos[p]} \rVert^2_2 \\
		& = \frac{2}{| \spatialNeighborhood_{\imgPos} | \cdot | \spatialNeighborhood_{\imgPos[j]} |} \sum_{\imgPos[q] \in \spatialNeighborhood_{\imgPos[j]}}  \sum_{\imgPos[p] \in \spatialNeighborhood_{\imgPos}} \lVert \dataPointHighDim_{\imgPos[q]} - \dataPointHighDim_{\imgPos[p]} \rVert^2_2
	\end{split}
	\label{eq:sup:sumofdiffs}
\end{equation}

Weighted versions analogous to \autoref{sec:method:weights}:

\begin{align}
	d^{PC}_{Haus}(\featureMat_{\imgPos}, \featureMat_{\imgPos[j]})
	& = \max \biggl\{ \biggr. %
	\max_{\imgPos[q] \in \spatialNeighborhood_{\imgPos}} \left( w_{\imgPos[q]} \min_{\imgPos[p] \in \spatialNeighborhood_{\imgPos[j]}} \lVert \dataPointHighDim_{\imgPos[q]} - \dataPointHighDim_{\imgPos[p]} \rVert^2_2\ \right), 
	\max_{\imgPos[q] \in \spatialNeighborhood_{\imgPos[j]}} \left( w_{\imgPos[q]} \min_{\imgPos[p] \in \spatialNeighborhood_{\imgPos}} \lVert \dataPointHighDim_{\imgPos[q]} - \dataPointHighDim_{\imgPos[p]} \rVert^2_2 \right)  \biggl. \biggr\} %
	\label{eq:sup:hausdorff_weighted} \\
	d^{PC}_{HM}(\featureMat_{\imgPos}, \featureMat_{\imgPos[j]})
	&= \frac{1}{2} \left(
	\operatorname*{median}_{\imgPos[q] \in \spatialNeighborhood_{\imgPos}} \left( w_{\imgPos[q]} \min_{\imgPos[p] \in \spatialNeighborhood_{\imgPos[j]}} \lVert \dataPointHighDim_{\imgPos[q]} - \dataPointHighDim_{\imgPos[p]} \rVert^2_2\ \right) +\ 
	\operatorname*{median}_{\imgPos[q] \in \spatialNeighborhood_{\imgPos[j]}} \left( w_{\imgPos[q]} \min_{\imgPos[p] \in \spatialNeighborhood_{\imgPos}} \lVert \dataPointHighDim_{\imgPos[q]} - \dataPointHighDim_{\imgPos[p]} \rVert^2_2 \right) \right) 
	\label{eq:sup:haus_med_weighted} \\
	d^{PC}_{SSD}(\featureMat_{\imgPos}, \featureMat_{\imgPos[j]}) 
	& = \frac{2}{| \spatialNeighborhood_{\imgPos} | \cdot | \spatialNeighborhood_{\imgPos[j]} |} \sum_{\imgPos[q] \in \spatialNeighborhood_{\imgPos[j]}}  \sum_{\imgPos[p] \in \spatialNeighborhood_{\imgPos}} ( w_{\imgPos[q]} + w_{\imgPos[p]} ) \lVert \dataPointHighDim_{\imgPos[q]} - \dataPointHighDim_{\imgPos[p]} \rVert^2_2
	\label{eq:sup:sumofdiffs_weighted}
\end{align}

\begin{figure*}[ht]
    \centering
    \begin{subfigure}[b]{0.4\textwidth}
        \includegraphics[width=\textwidth]{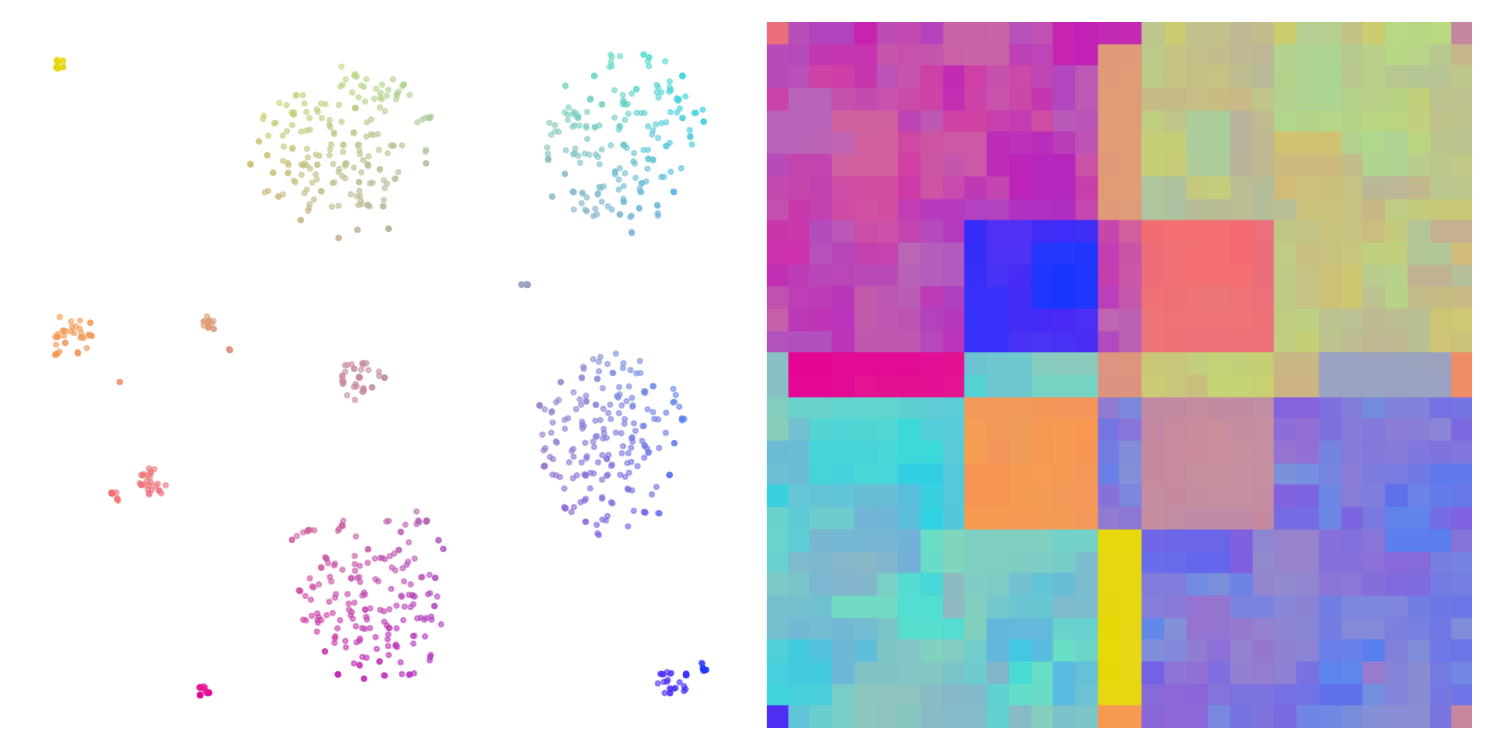}
        \caption{Hausdorff}
        \label{fig:supp:HausdorffFamiliy:Haus}        
    \end{subfigure}
    \qquad
    \begin{subfigure}[b]{0.4\textwidth}
        \includegraphics[width=\textwidth]{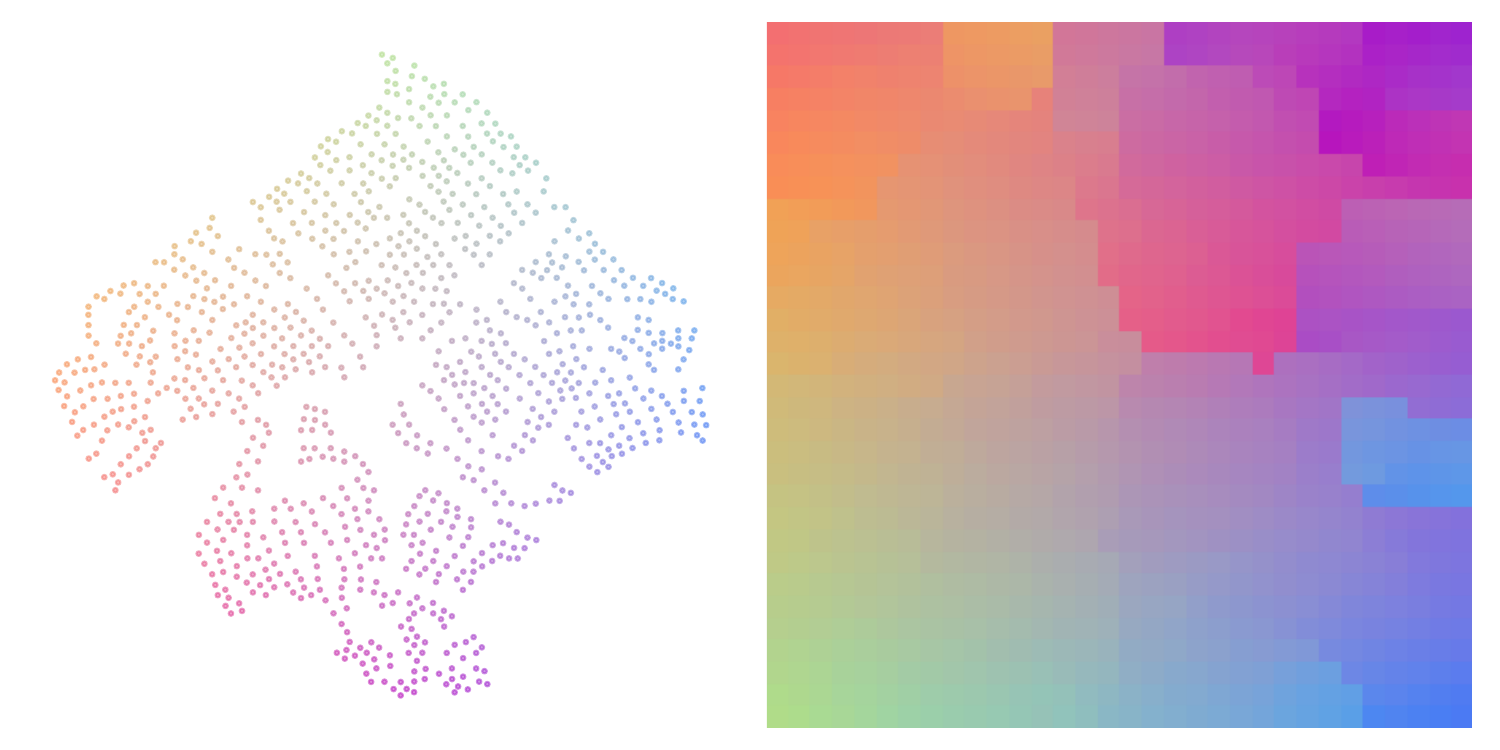}
        \caption{Hausdorff Median}
        \label{fig:supp:HausdorffFamiliy:HausMed}        
    \end{subfigure}

    \begin{subfigure}[b]{0.4\textwidth}
        \includegraphics[width=\textwidth]{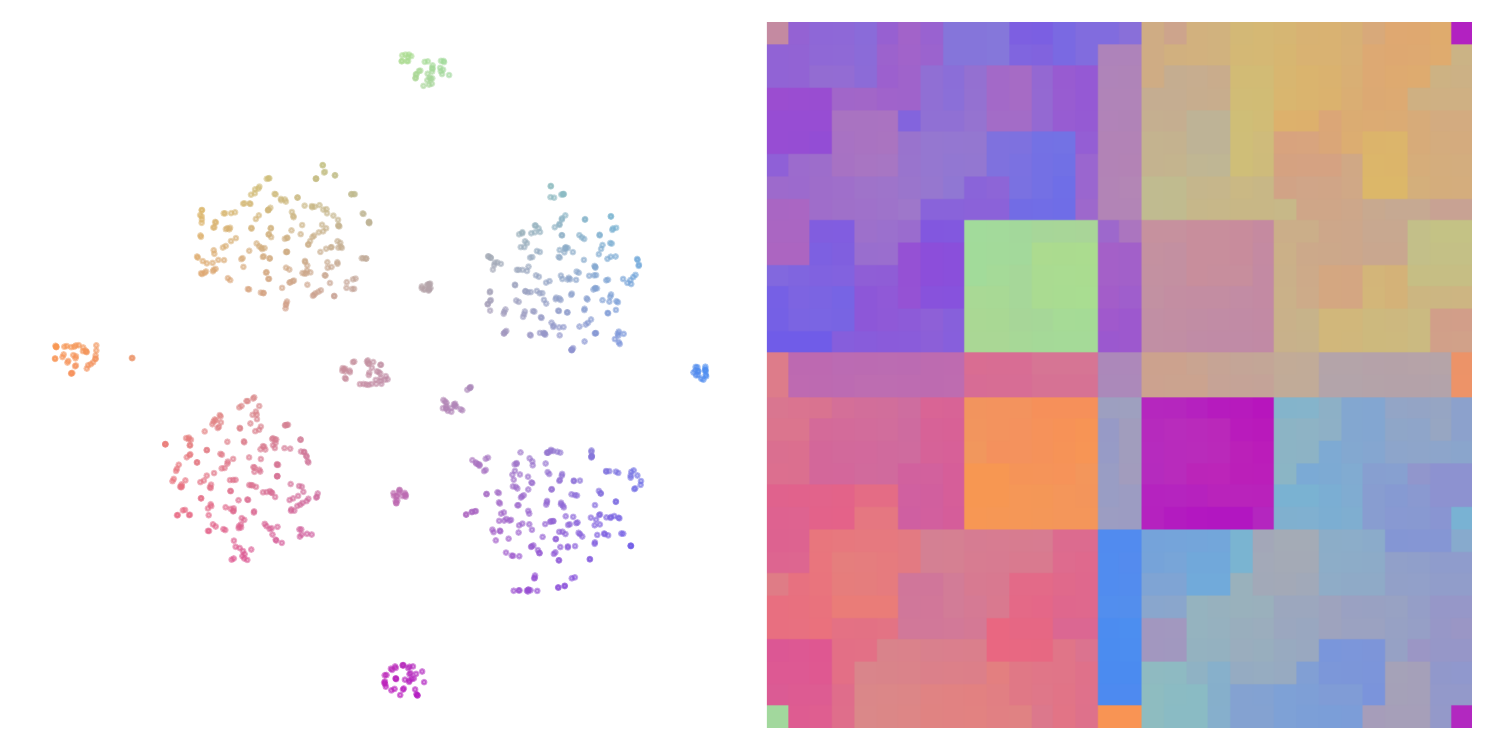}
        \caption{Chamfer}
        \label{fig:supp:HausdorffFamiliy:Chamfer}        
    \end{subfigure}
    \qquad
    \begin{subfigure}[b]{0.4\textwidth}
        \includegraphics[width=\textwidth]{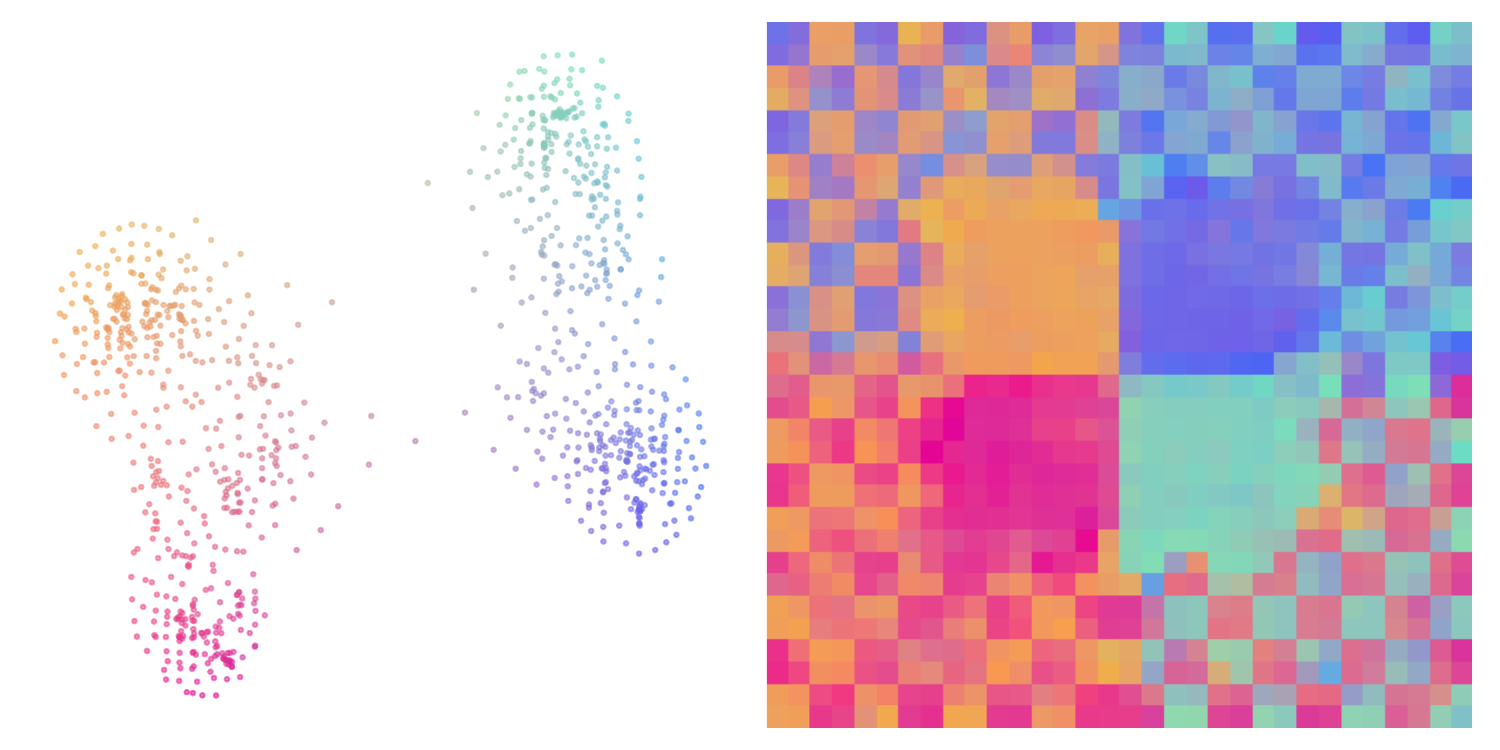}
        \caption{SSD}
        \label{fig:supp:HausdorffFamiliy:SSD}        
    \end{subfigure}

    \caption{All spatially informed embeddings are computed with a 3x3 neighbourhood.}
    \label{fig:supp:HausdorffFamiliy}
\end{figure*}

\begin{figure*}[ht]
    \centering
    \includegraphics[width=0.5\textwidth]{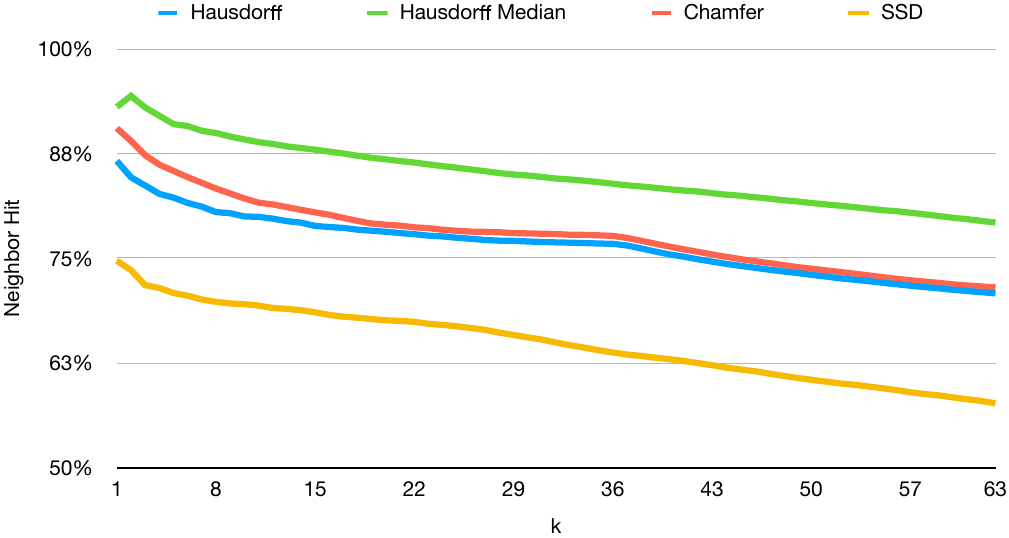}

    \caption{Nearest neighbor hit for the four embeddings in \autoref{fig:supp:HausdorffFamiliy}}
    \label{fig:supp:HausdorffFamilyQuant}
\end{figure*}

See \Autoref{fig:supp:HausdorffFamiliy, fig:supp:HausdorffFamilyQuant} for a comparison of these point cloud distances for the synthetic data set from the main paper.

\clearpage
\subsection*{S4: Varying neighborhood sizes and spatial weighting} \labelshort[S4]{supp:weighting} %

\begin{figure*}[!h]
    \centering

    \begin{subfigure}[b]{0.31\textwidth}
        \includegraphics[width=\textwidth]{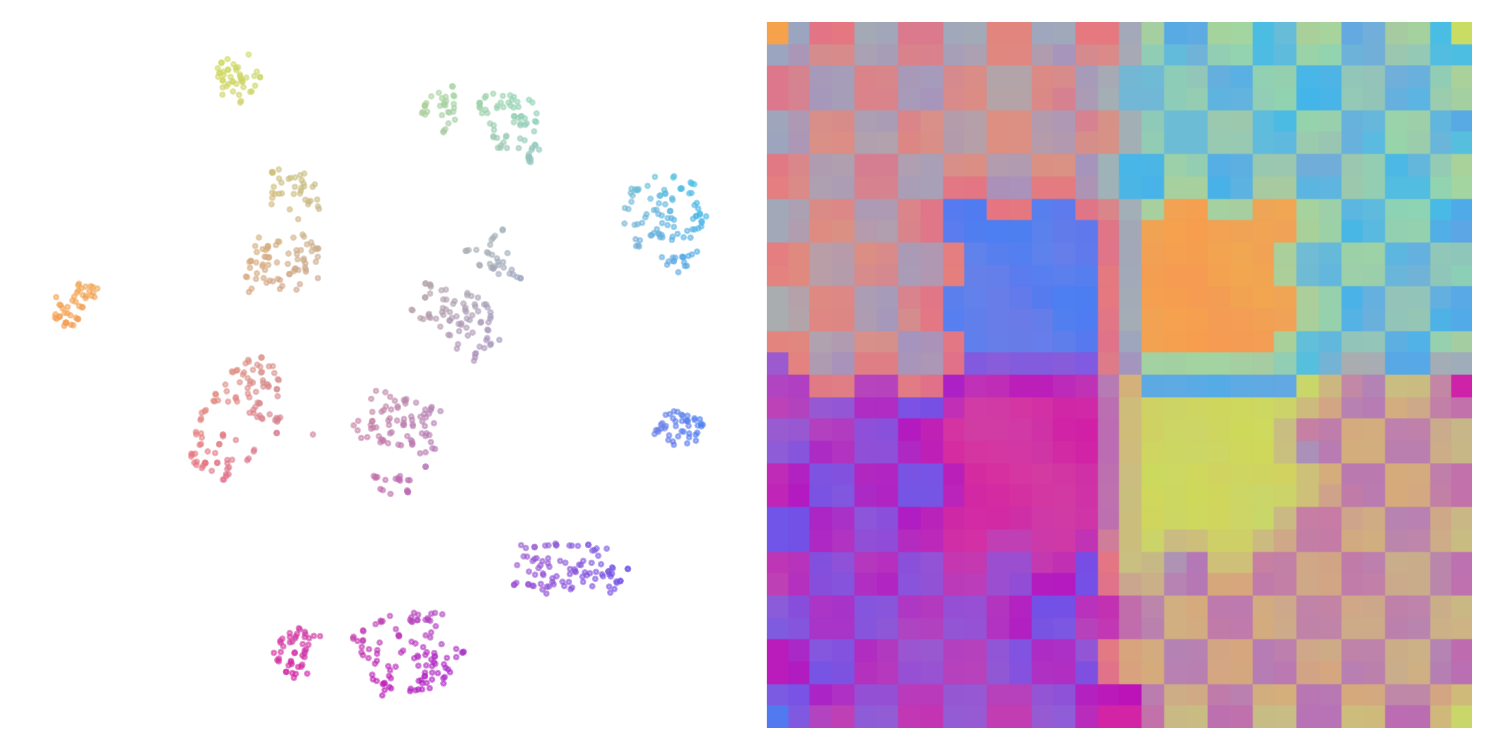}
        \caption{Local histograms}
        \label{fig:supp:3x3GW:qf}        
    \end{subfigure}
\hfill
    \begin{subfigure}[b]{0.31\textwidth}
        \includegraphics[width=\textwidth]{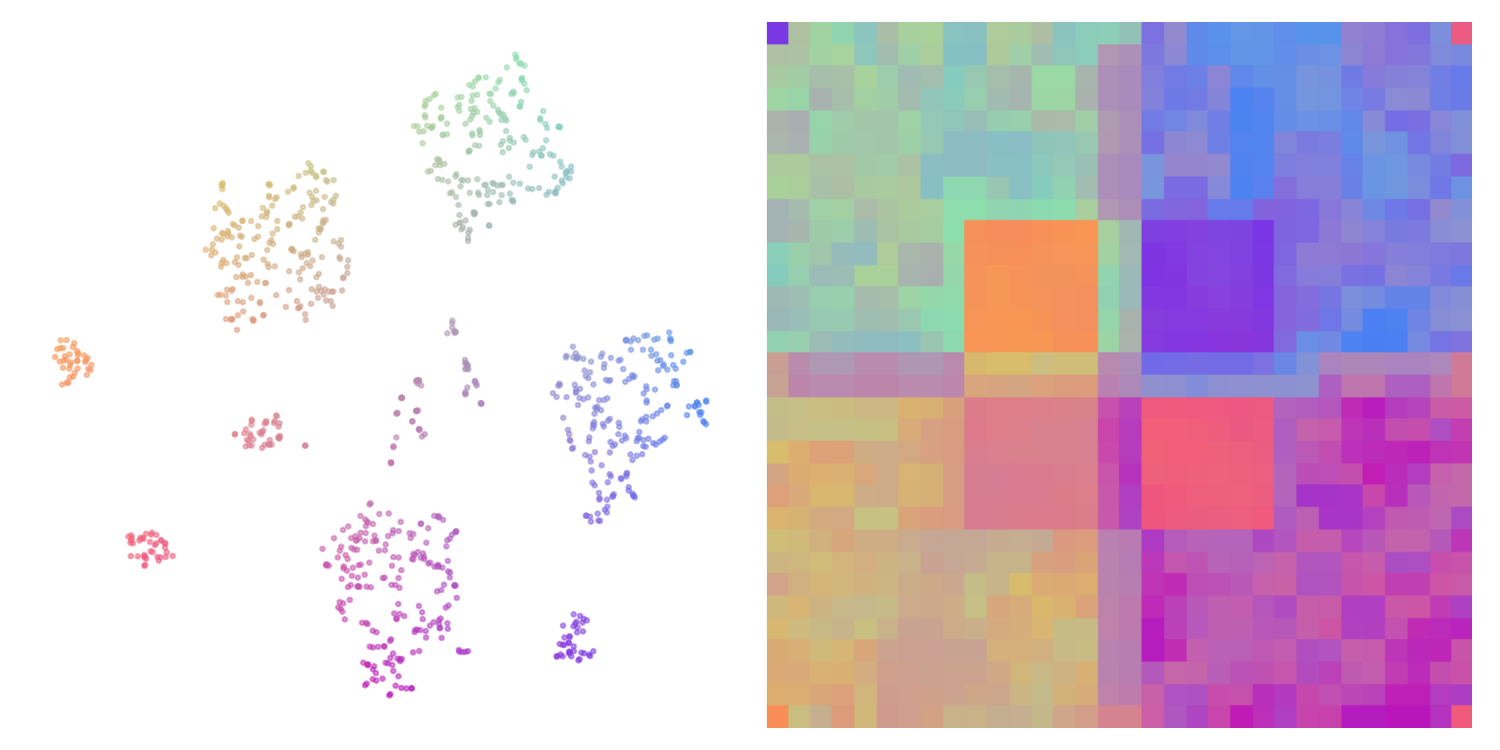}
        \caption{Covariance matrix and means}
        \label{fig:supp:3x3GW:cov}        
    \end{subfigure}
\hfill
    \begin{subfigure}[b]{0.31\textwidth}
        \includegraphics[width=\textwidth]{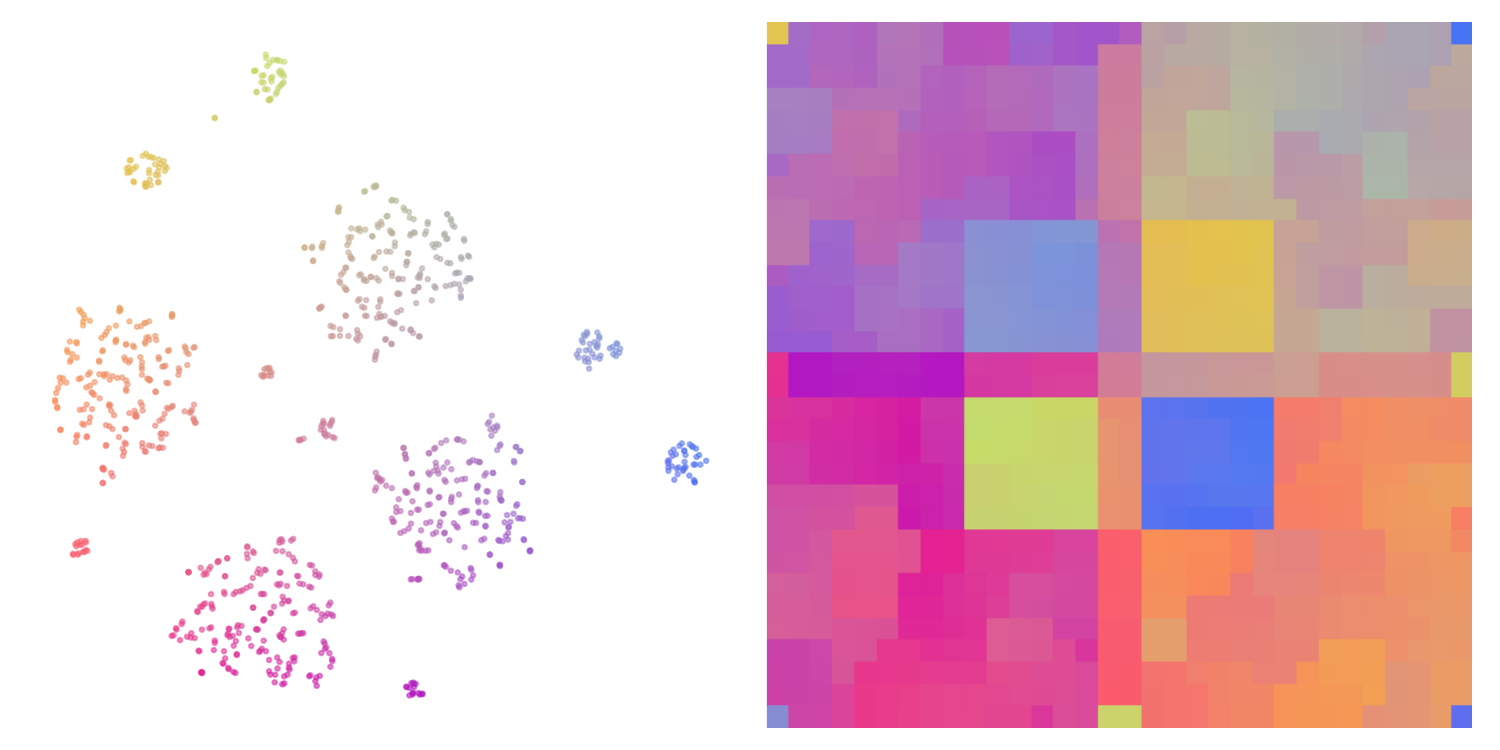}
        \caption{Chamfer point cloud distance}
        \label{fig:supp:3x3GW:pc}        
    \end{subfigure}
    
    \vspace{-5pt}
    \caption{All spatially informed embeddings are computed with a 3x3 neighbourhood and Gaussian weighting.}
    \label{fig:supp:3x3GW}
\end{figure*}

\begin{figure*}[!h]
    \centering

    \begin{subfigure}[b]{0.31\textwidth}
        \includegraphics[width=\textwidth]{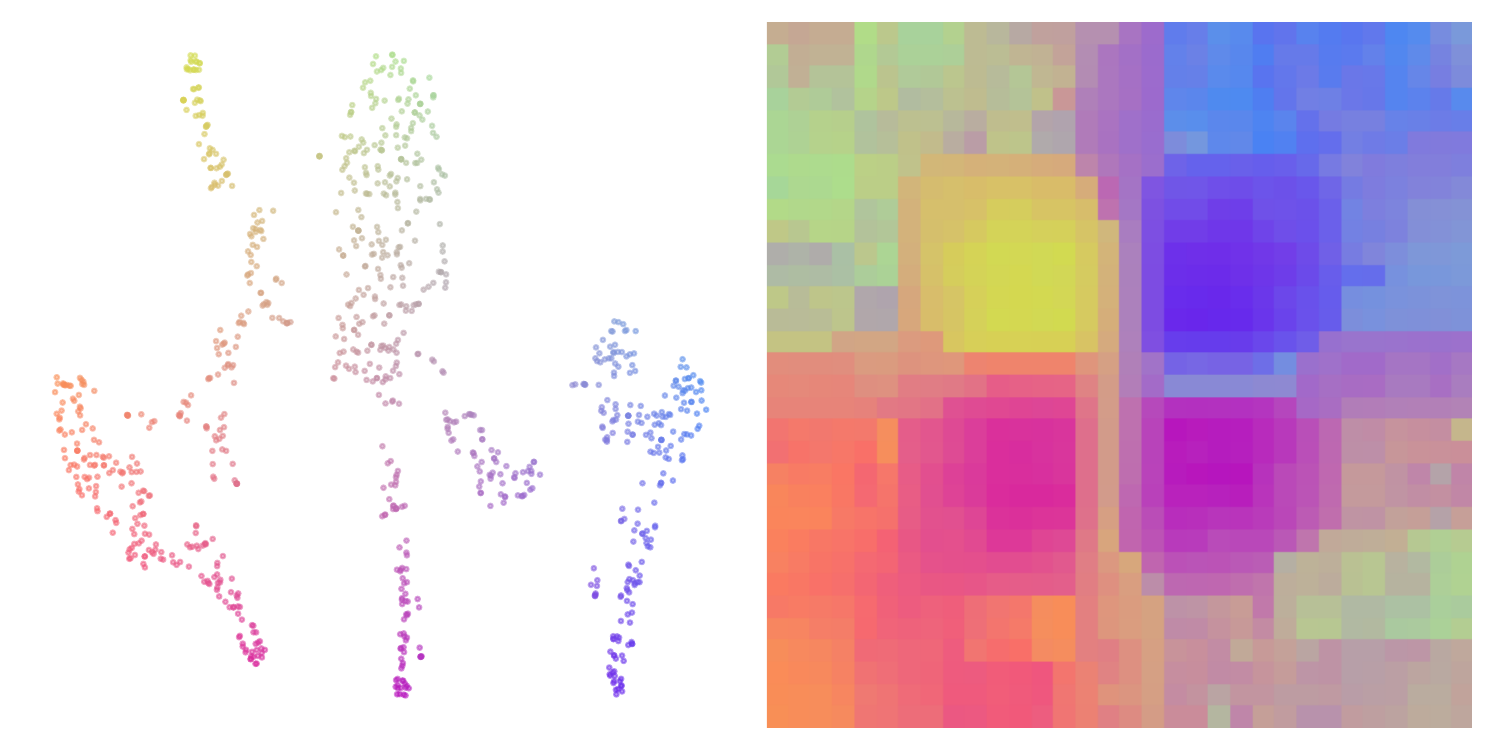}
        \caption{Local histograms}
        \label{fig:supp:5x5:qf}        
    \end{subfigure}
    \hfill
    \begin{subfigure}[b]{0.31\textwidth}
        \includegraphics[width=\textwidth]{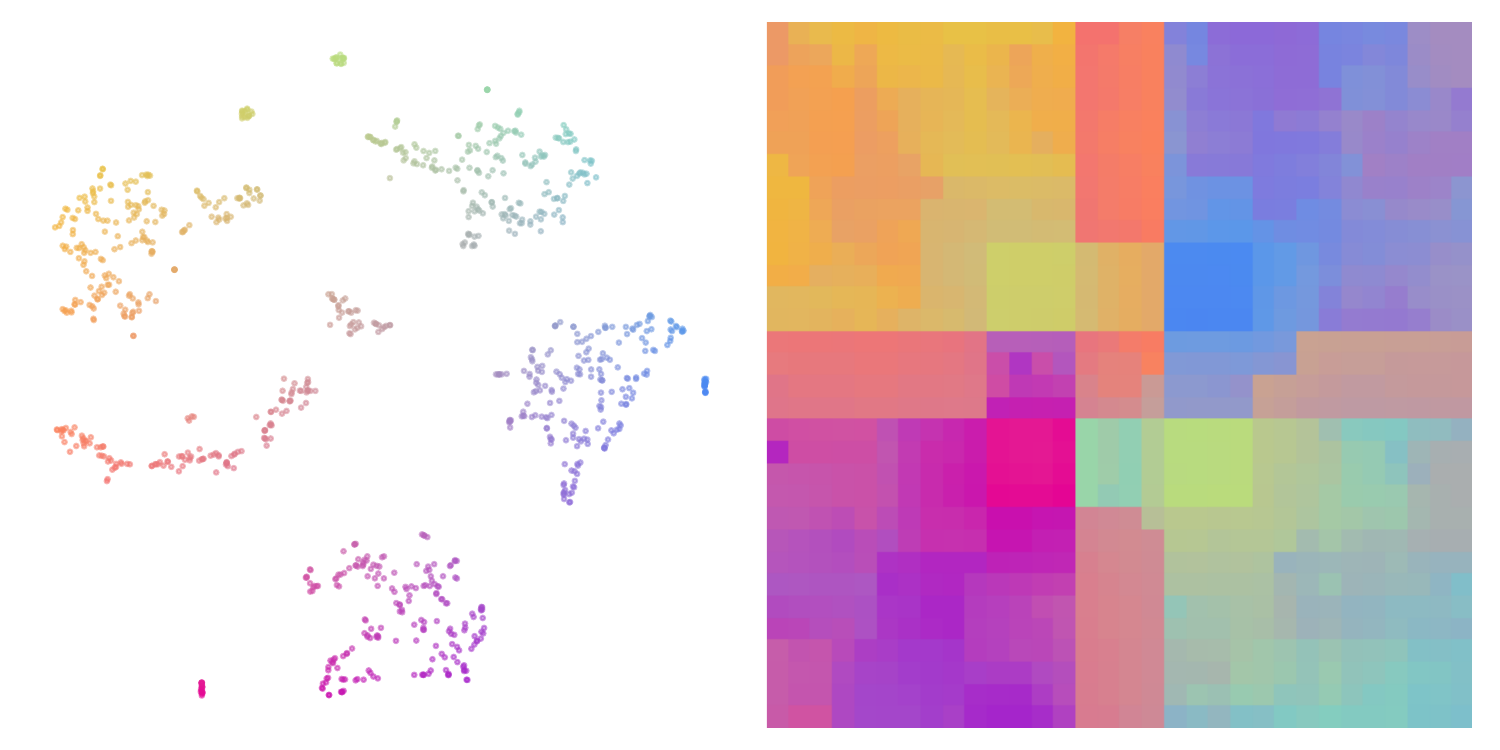}
        \caption{Covariance matrix and means}
        \label{fig:supp:5x5:cov}        
    \end{subfigure}
    \hfill
    \begin{subfigure}[b]{0.31\textwidth}
        \includegraphics[width=\textwidth]{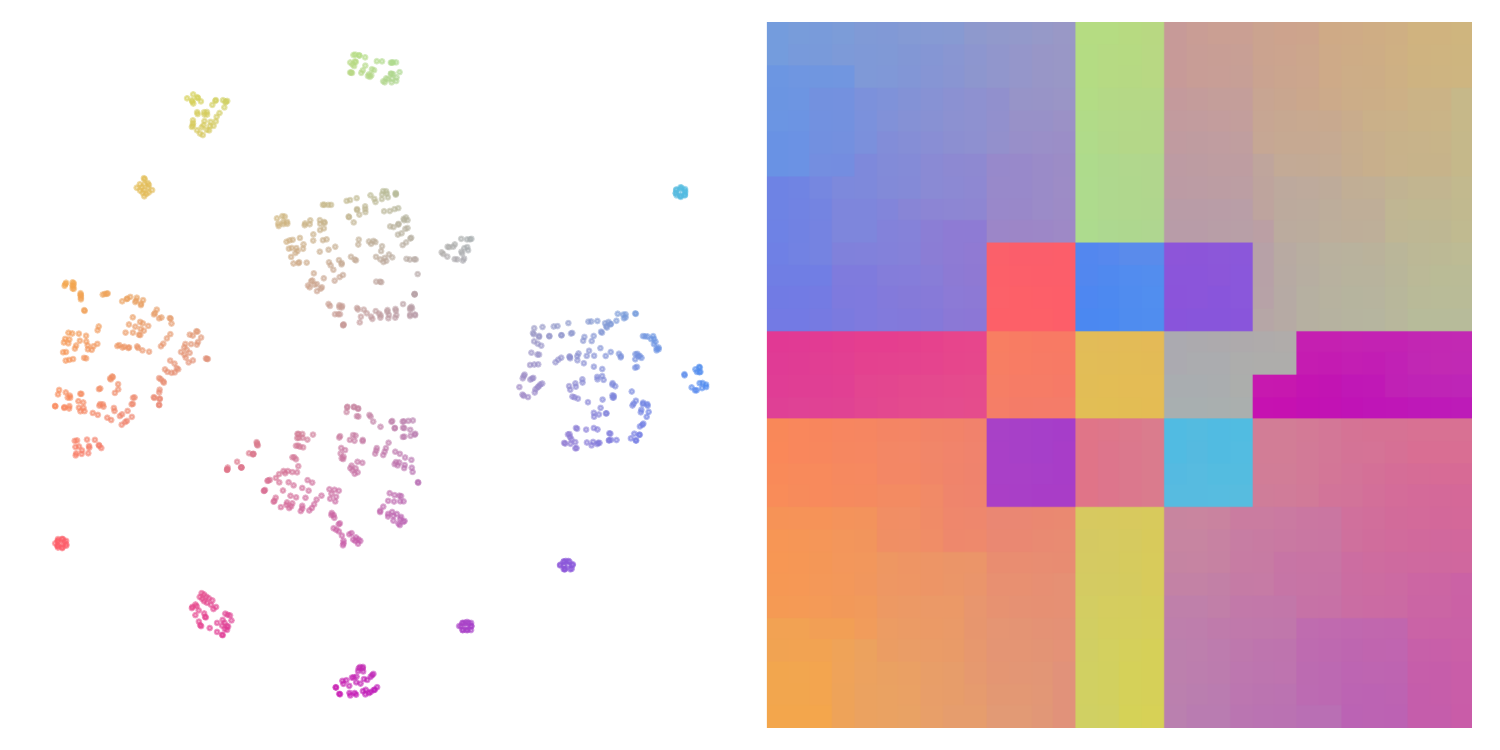}
        \caption{Chamfer point cloud distance}
        \label{fig:supp:5x5:pc}        
    \end{subfigure}

    \vspace{-5pt}
    \caption{All spatially informed embeddings are computed with a 5x5 neighbourhood.}
    \label{fig:supp:5x5}
\end{figure*}

\begin{figure*}[!h]
    \centering

    \begin{subfigure}[b]{0.31\textwidth}
        \includegraphics[width=\textwidth]{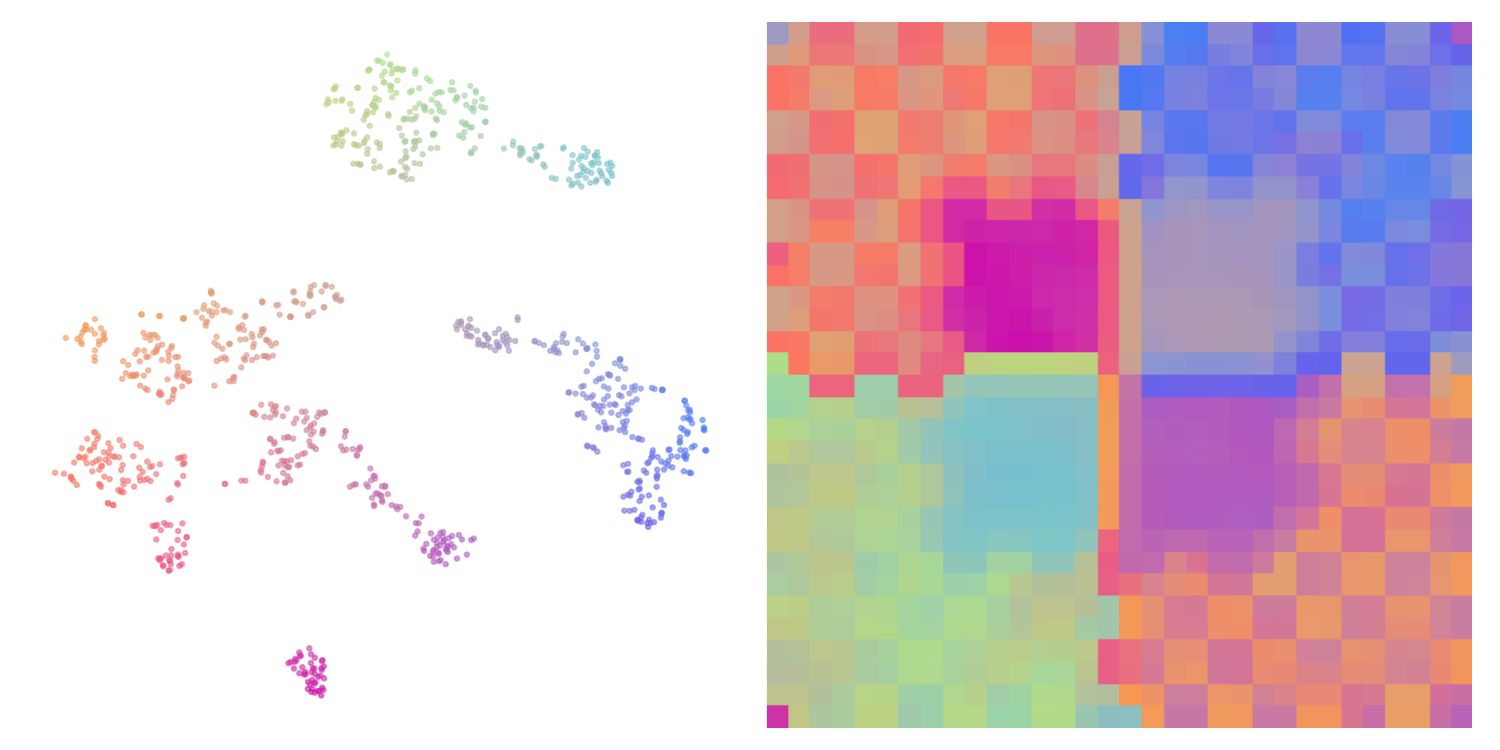}
        \caption{Local histograms}
        \label{fig:supp:5x5GW:qf}        
    \end{subfigure}
    \hfill
    \begin{subfigure}[b]{0.31\textwidth}
        \includegraphics[width=\textwidth]{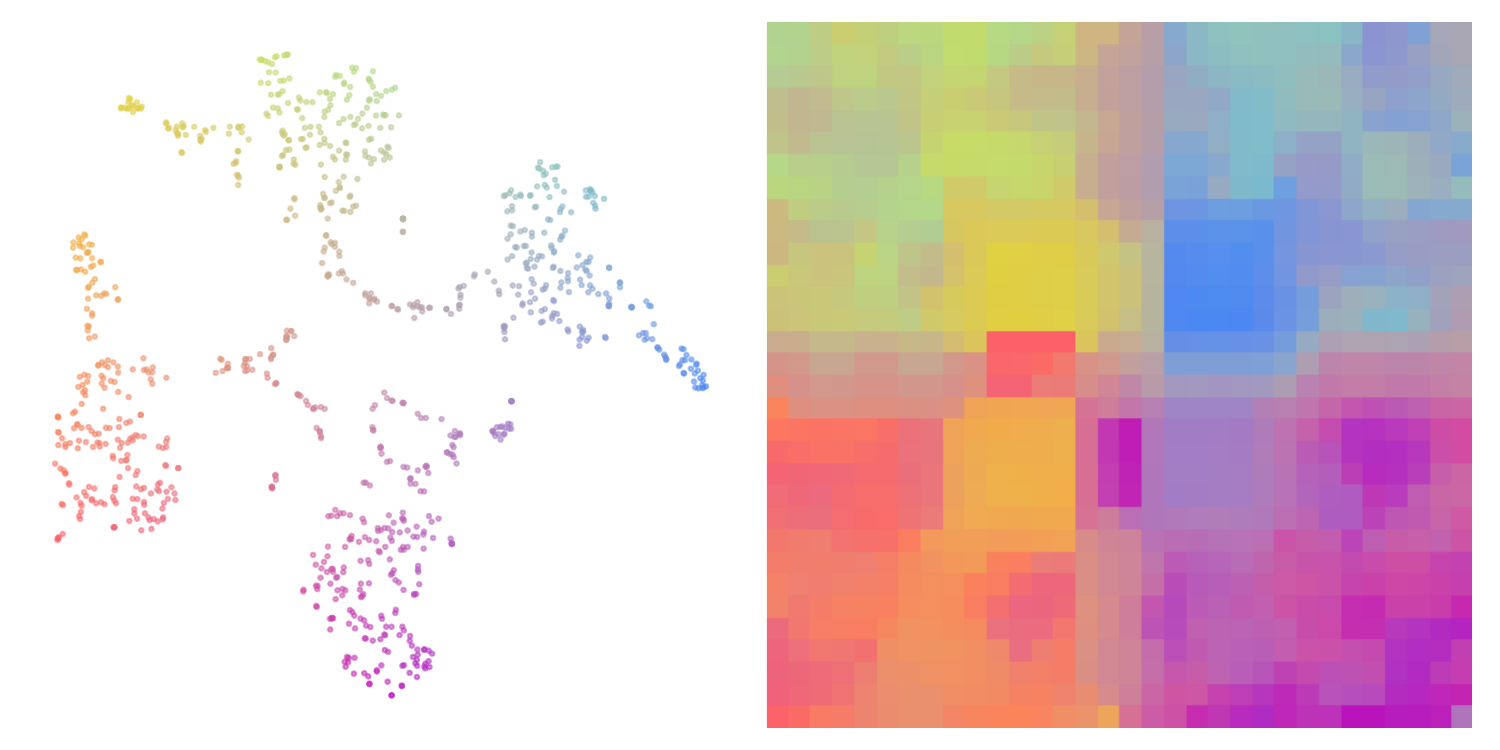}
        \caption{Covariance matrix and means}
        \label{fig:supp:5x5GW:cov}        
    \end{subfigure}
    \hfill
    \begin{subfigure}[b]{0.31\textwidth}
        \includegraphics[width=\textwidth]{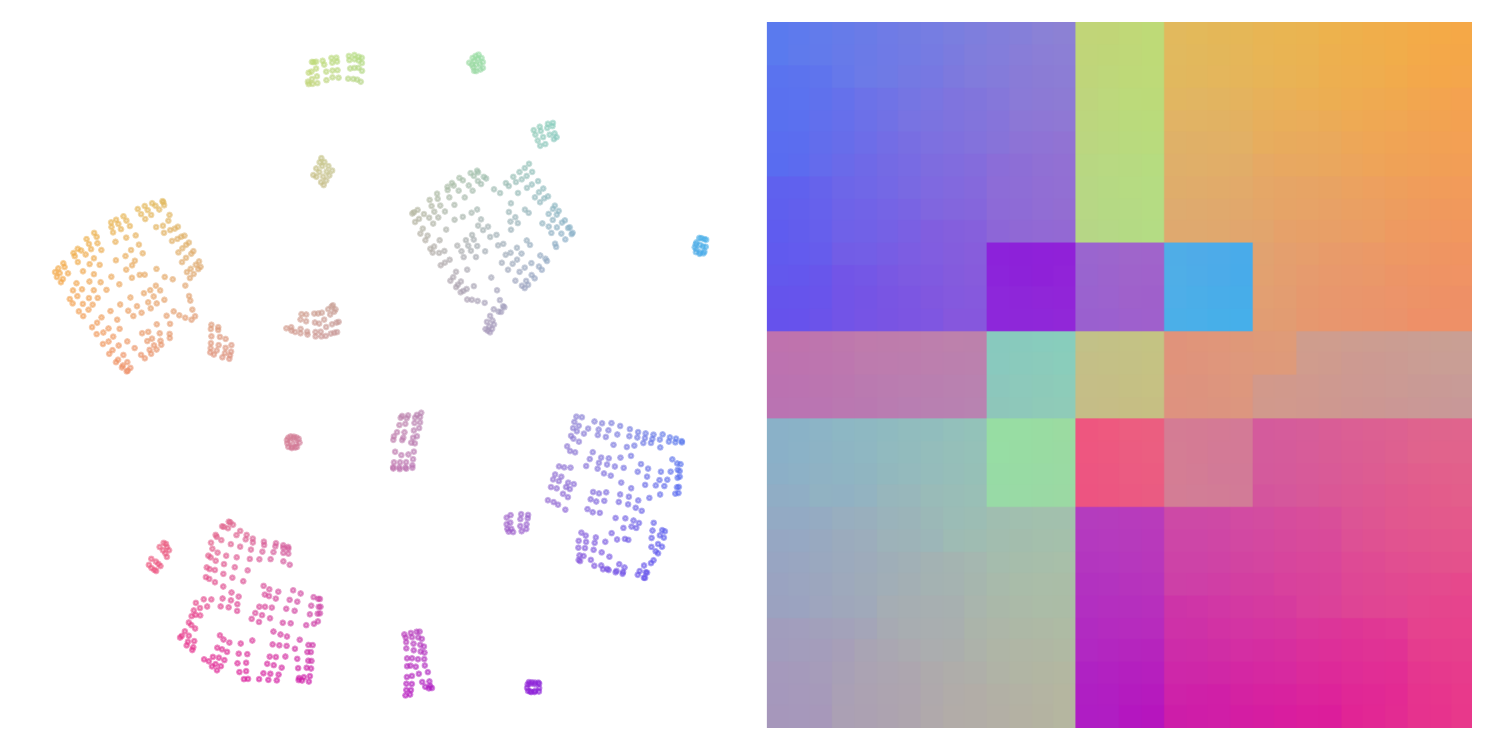}
        \caption{Chamfer point cloud distance}
        \label{fig:supp:5x5GW:pc}        
    \end{subfigure}

    \vspace{-5pt}
    \caption{All spatially informed embeddings are computed with a 5x5 neighbourhood and Gaussian weighting.}
    \label{fig:supp:5x5GW}
\end{figure*}

\begin{figure*}[!h]
    \centering

    \begin{subfigure}[b]{0.31\textwidth}
        \includegraphics[width=\textwidth]{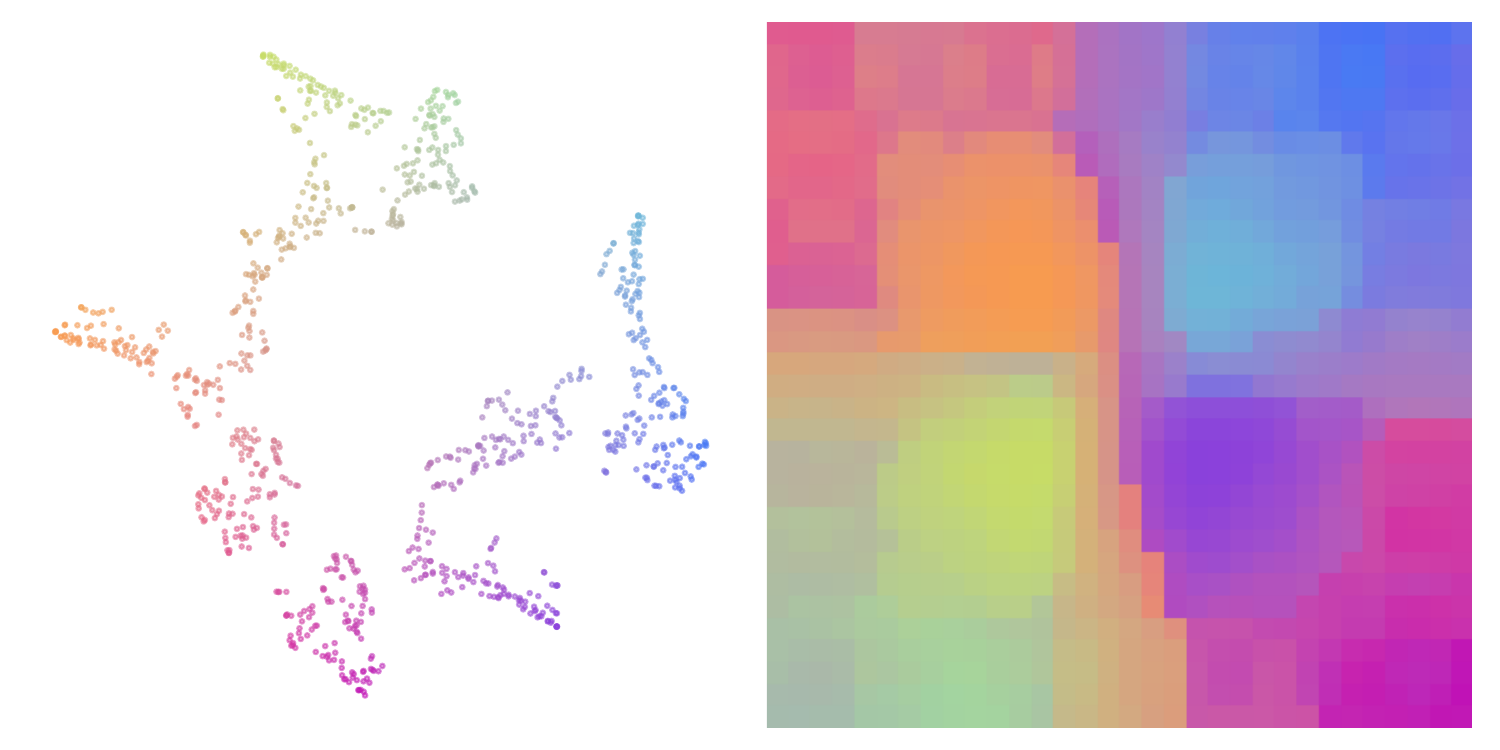}
        \caption{Local histograms}
        \label{fig:supp:7x7:f}        
    \end{subfigure}
    \hfill
    \begin{subfigure}[b]{0.31\textwidth}
        \includegraphics[width=\textwidth]{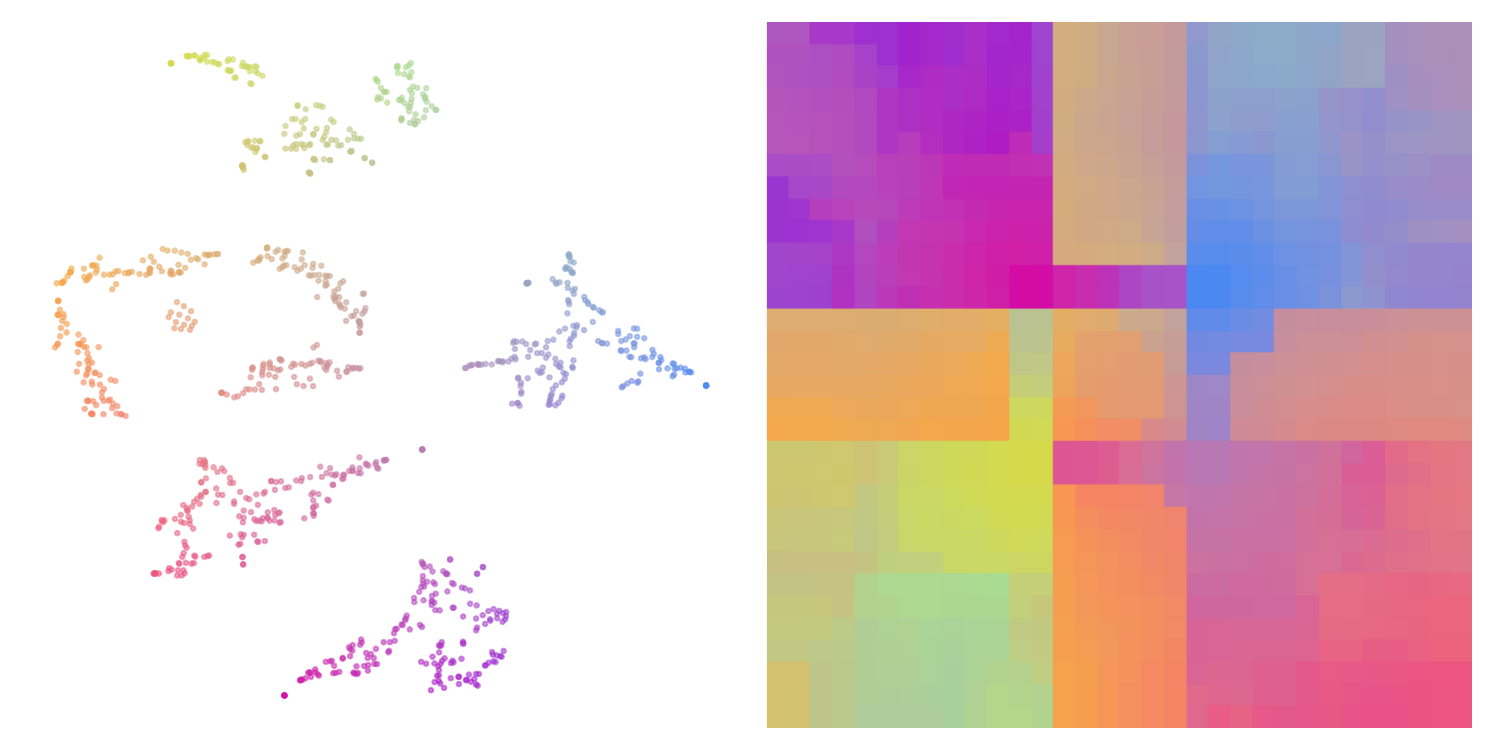}
        \caption{Covariance matrix and means}
        \label{fig:supp:7x7:cov}        
    \end{subfigure}
    \hfill
    \begin{subfigure}[b]{0.31\textwidth}
        \includegraphics[width=\textwidth]{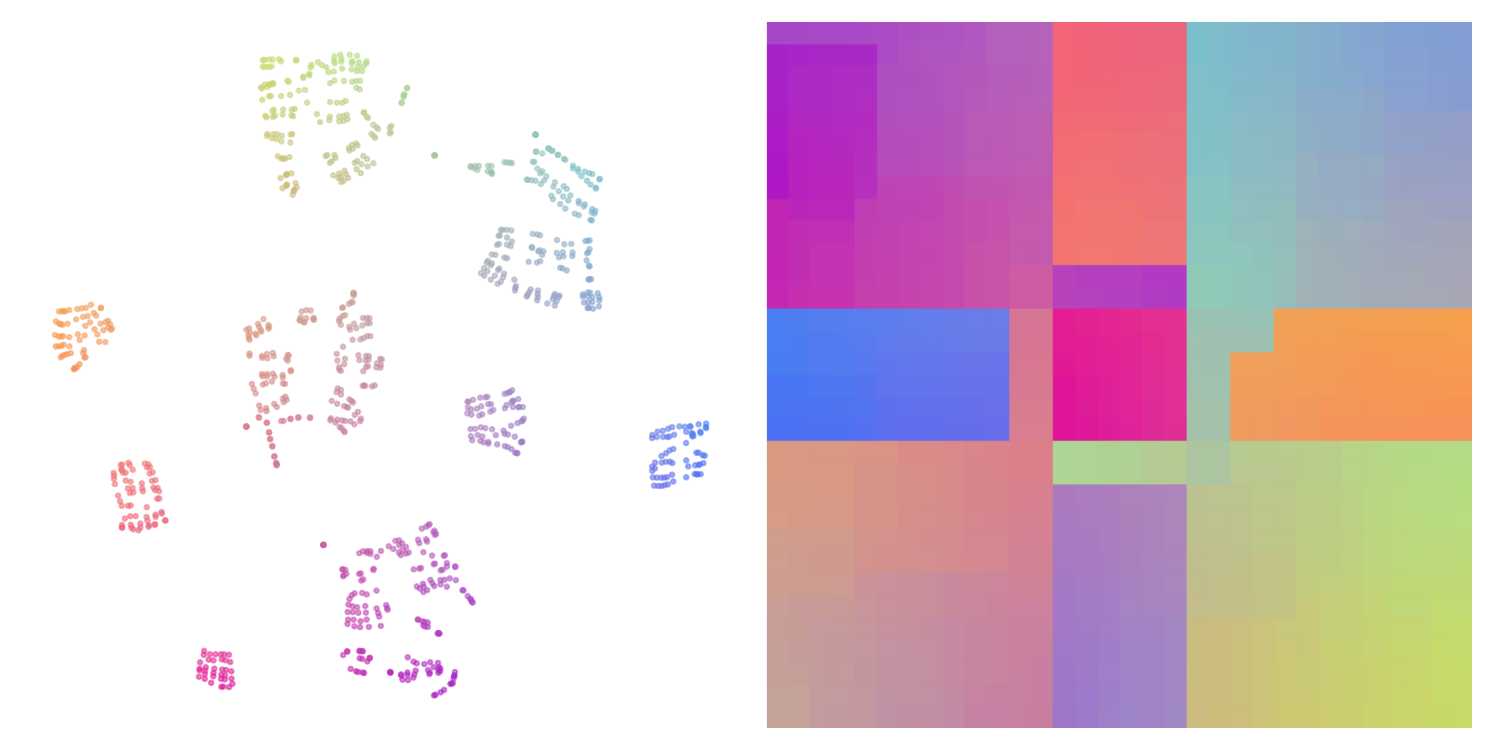}
        \caption{Chamfer point cloud distance}
        \label{fig:supp:7x7:pc}        
    \end{subfigure}

    \vspace{-5pt}
    \caption{All spatially informed embeddings are computed with a 7x7 neighbourhood.}
    \label{fig:supp:7x7}
\end{figure*}

\begin{figure*}[!h]
    \centering

    \begin{subfigure}[b]{0.31\textwidth}
        \includegraphics[width=\textwidth]{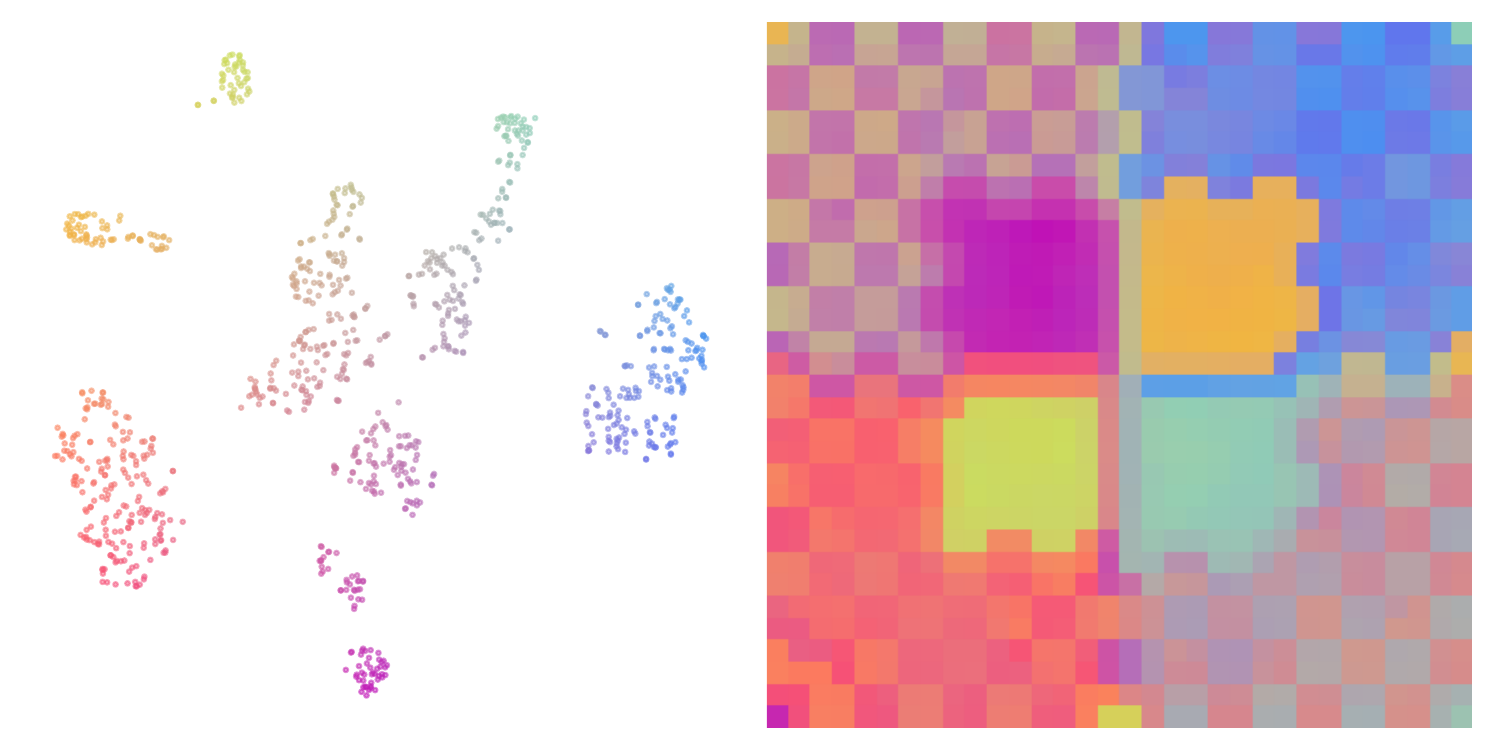}
        \caption{Local histograms}
        \label{fig:supp:7x7GW:qf}        
    \end{subfigure}
    \hfill
    \begin{subfigure}[b]{0.31\textwidth}
        \includegraphics[width=\textwidth]{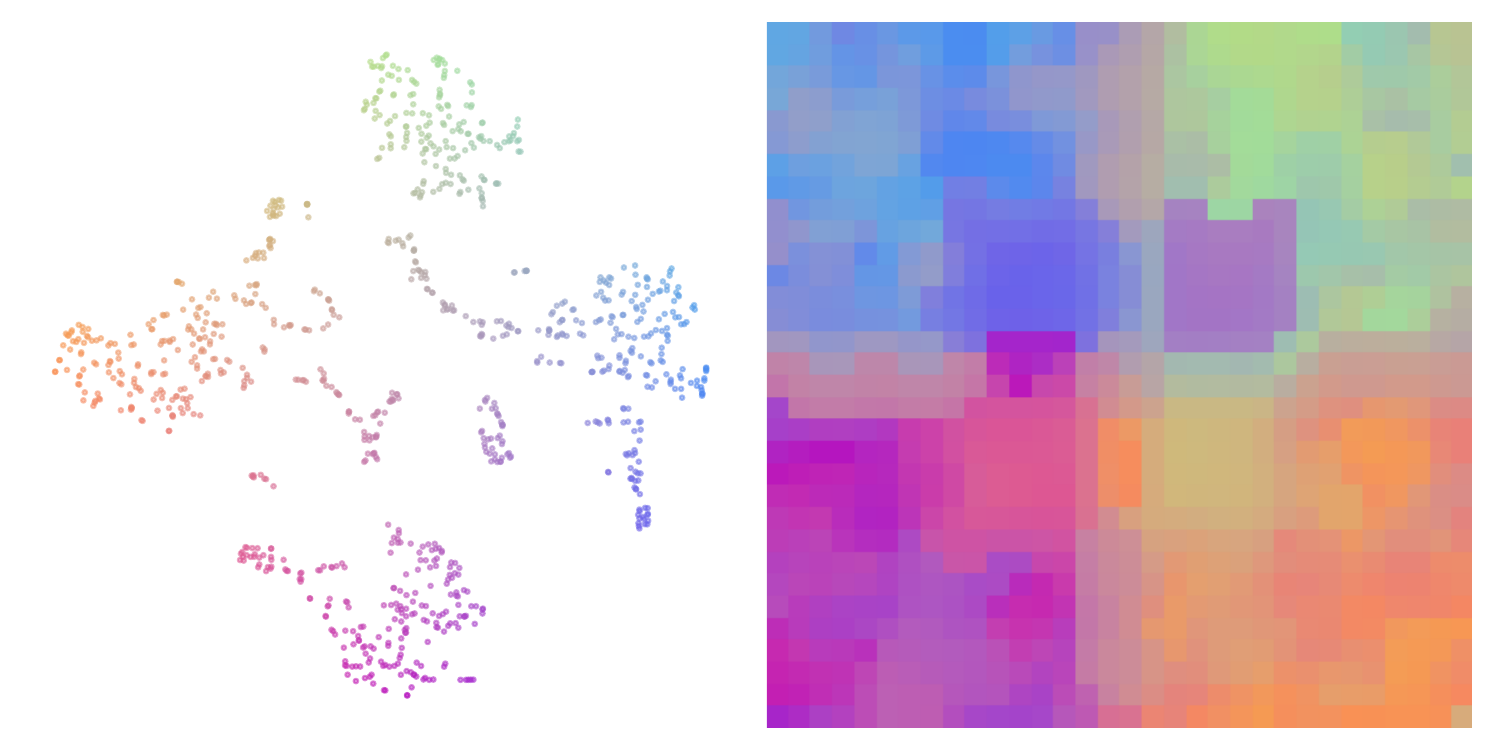}
        \caption{Covariance matrix and means}
        \label{fig:supp:7x7GW:cov}        
    \end{subfigure}
    \hfill
    \begin{subfigure}[b]{0.31\textwidth}
        \includegraphics[width=\textwidth]{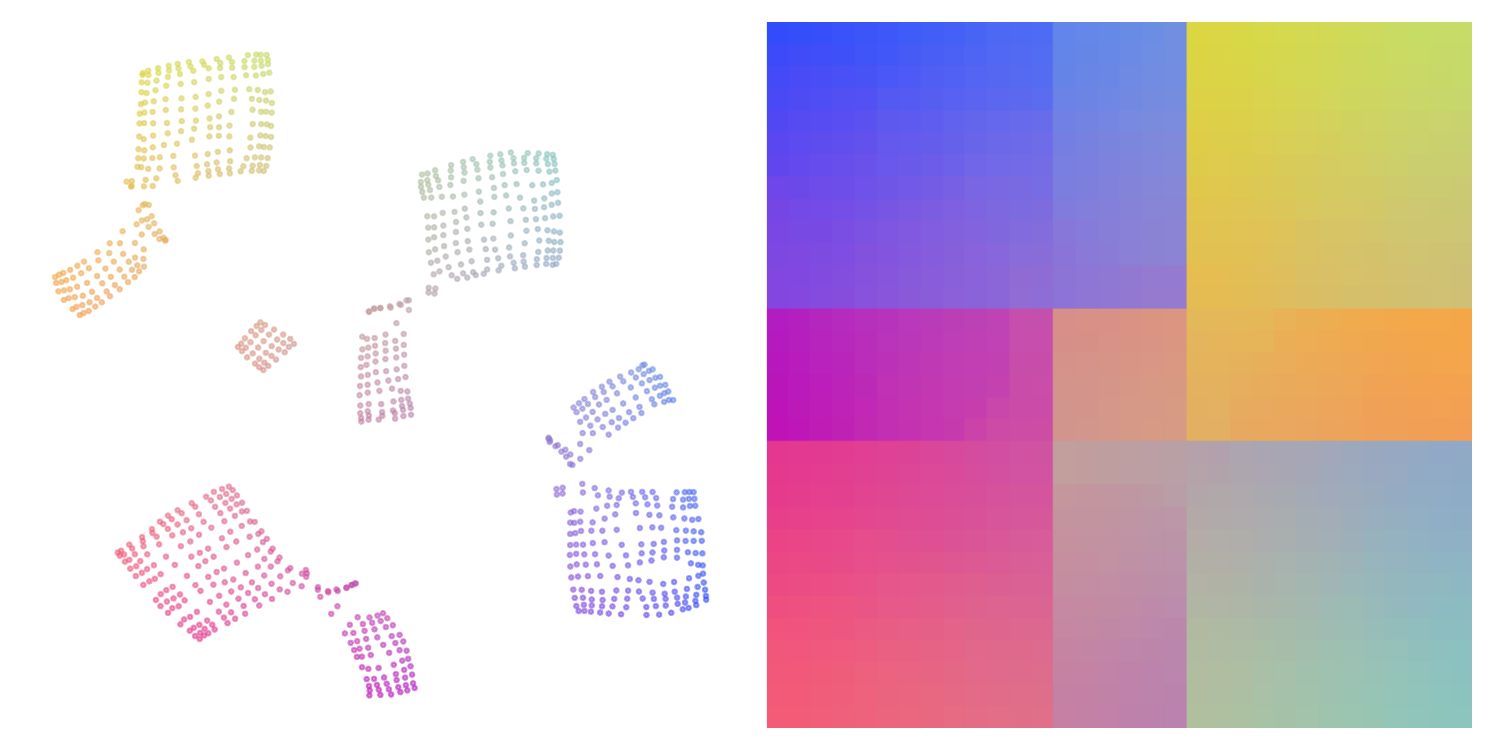}
        \caption{Chamfer point cloud distance}
        \label{fig:supp:7x7GW:pc}        
    \end{subfigure}

    \vspace{-5pt}
    \caption{All spatially informed embeddings are computed with a 7x7 neighbourhood and Gaussian weighting.}
    \label{fig:supp:7x7GW}
\end{figure*}

See~\Autoref{fig:supp:3x3GW, fig:supp:5x5, fig:supp:5x5GW, fig:supp:7x7, fig:supp:7x7GW} for an overview of of the effect of different neighborhood sizes for the synthetic data set from the main paper.

\begin{figure*}[!h]
    \centering

    \begin{subfigure}[b]{0.31\textwidth}
        \includegraphics[width=\textwidth]{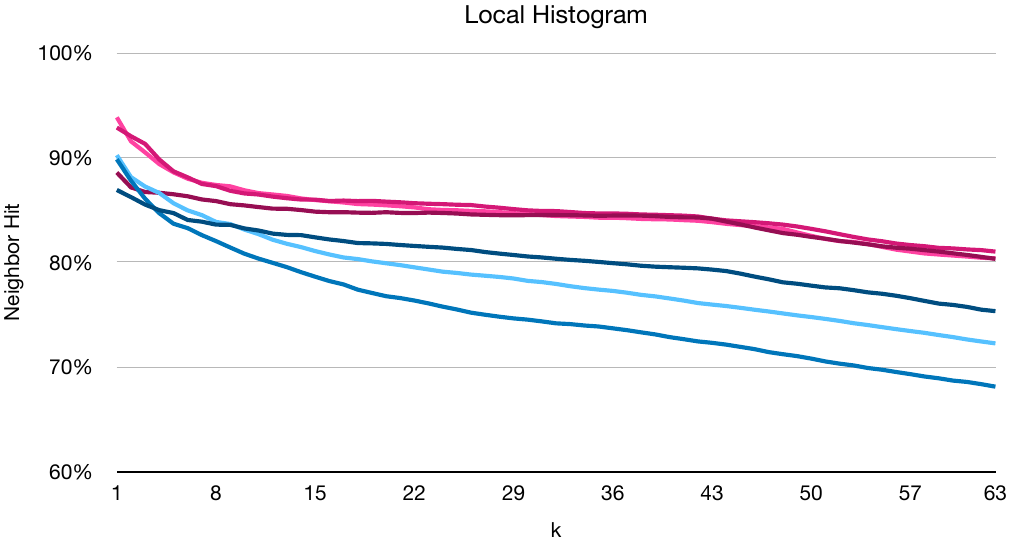}
        \caption{Local histograms}
        \label{fig:supp:neighborhit:f}        
    \end{subfigure}
    \hfill
    \begin{subfigure}[b]{0.31\textwidth}
        \includegraphics[width=\textwidth]{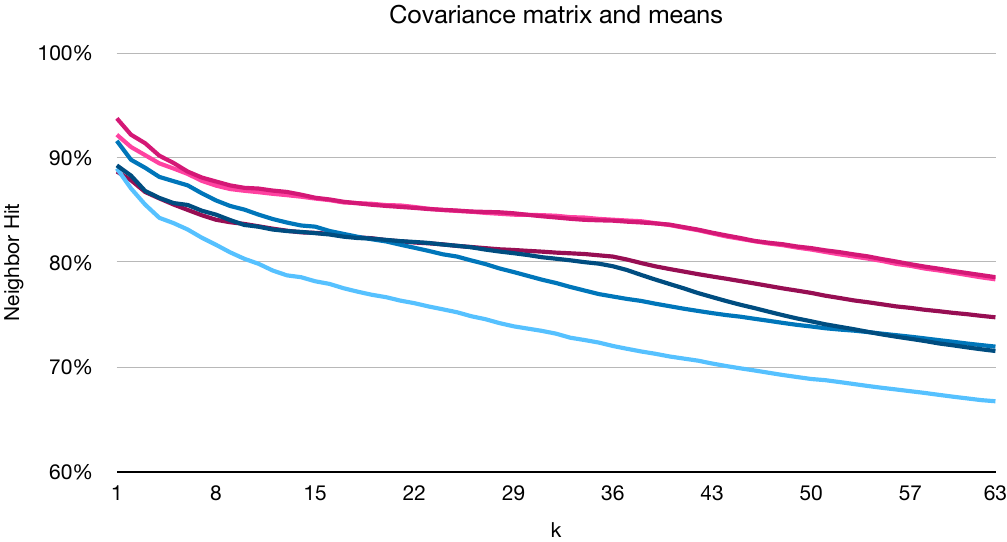}
        \caption{Covariance matrix and means}
        \label{fig:supp:neighborhit:cov}        
    \end{subfigure}
    \hfill
    \begin{subfigure}[b]{0.31\textwidth}
        \includegraphics[width=\textwidth]{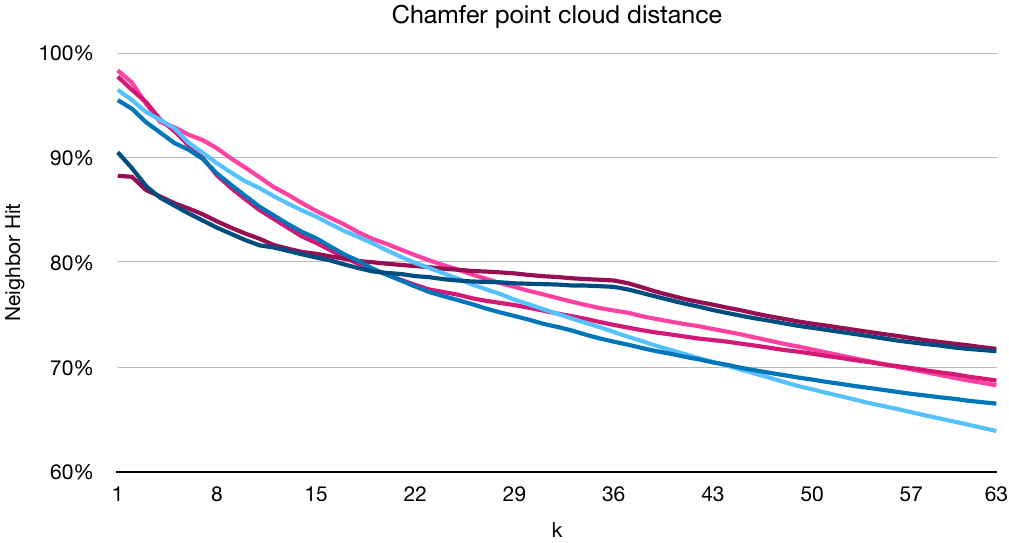}
        \caption{Chamfer point cloud distance}
        \label{fig:supp:neighborhit:pc}        
    \end{subfigure}

    \vspace{-5pt}
    \caption{Neighbor hit values for the local histogram (\autoref{fig:ArtExamplesOverview:qf}), covariance (\autoref{fig:ArtExamplesOverview:cov}), and point cloud (\autoref{fig:ArtExamplesOverview:pc}) -based embeddings and their different neighborhood size versions (\Autoref{fig:supp:3x3GW, fig:supp:5x5, fig:supp:5x5GW, fig:supp:7x7, fig:supp:7x7GW}). \textcolor{plotcolorblue1}{\textbf{\textemdash}} 3x3 neighborhood,
    \textcolor{plotcolorblue2}{\textbf{\textemdash}} 5x5 neighborhood, 
    \textcolor{plotcolorblue3}{\textbf{\textemdash}} 7x7 neighborhood, 
    \textcolor{plotcolorpink1}{\textbf{\textemdash}} 3x3 neighborhood Gaussian weighted, 
    \textcolor{plotcolorpink2}{\textbf{\textemdash}} 5x5 neighborhood Gaussian weighted, 
    \textcolor{plotcolorpink3}{\textbf{\textemdash}} 7x7 neighborhood Gaussian weighted.}
    \label{fig:supp:neighborhit}
\end{figure*}

\FloatBarrier
\subsection*{S5: Computation time evaluation} 
\begin{figure*}[ht]
    \centering
    \includegraphics[width=1\textwidth]{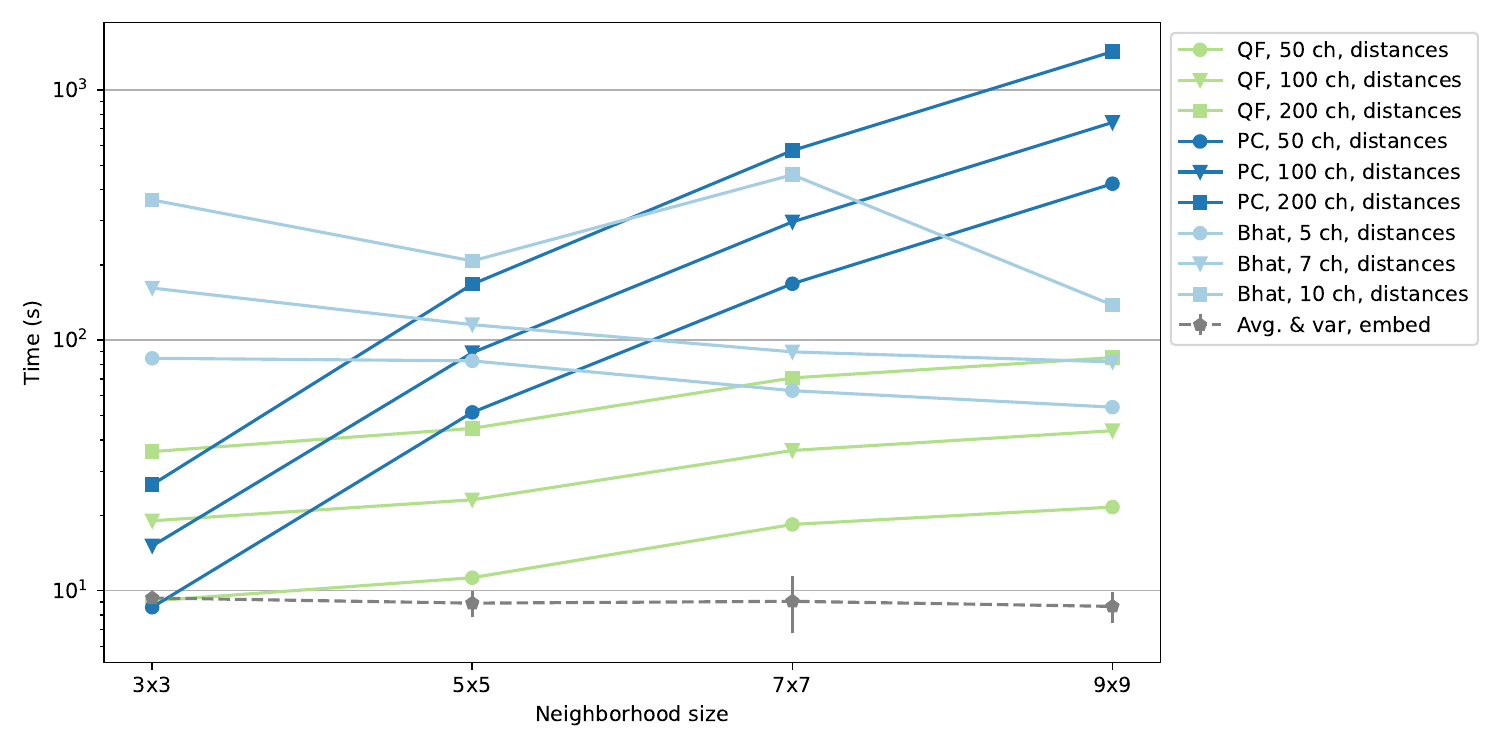}

    \caption{Computation time of the distance computation for varying neighborhood sizes: local histogram comparison with the quadratic form distance (QF), covariance matrix feature comparison with the Bhattacharyya distance (Bat) and the Chamfer point cloud distance (PC). The Indian Pines data set with $21.025$ data points and 200 channels was used for computation. The same random channel subsets were used for runs with less than 200 channels. The Bhattacharyya distance is listed for fewer channels since its runtime grows impractically large for higher channel counts, as shown in \autoref{fig:supp:num_channels}. As mentioned in the main paper, when using the QF distance we use the Rice rule to set the number of histogram bins. This results in 5, 6, 8 and 9 bins for the various neighborhood sizes respectively. The embedding time is not influenced by the distance metric and shown as an average of all measurements with variance bars.}
    \label{fig:supp:neighhorhood_size}
\end{figure*}

\begin{figure*}[ht]
    \centering
    \includegraphics[width=1\textwidth]{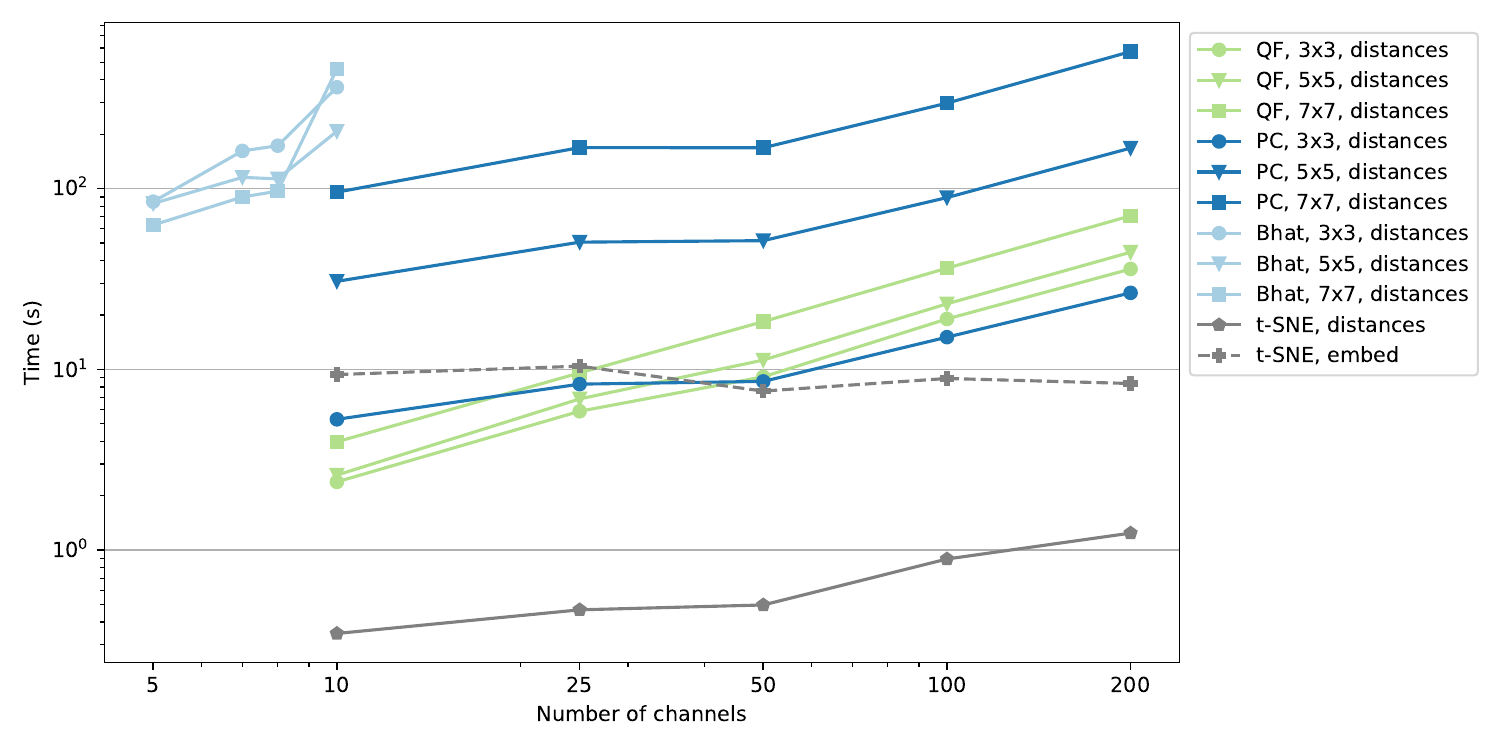}

    \caption{Computation time of the distance computation for channel numbers: local histogram comparison with the quadratic form distance (QF), covariance matrix feature comparison with the Bhattacharyya distance (Bat) and the Chamfer point cloud distance (PC). The Indian Pines data set with $21.025$ data points and 200 channels was used for computation. The same random channel subsets were used for runs with less than 200 channels. The Bhattacharyya distance is listed for fewer channels since its runtime grows impractically large for higher channel counts. The histogram bin number for the QF distance is set as described in \autoref{fig:supp:neighhorhood_size}. The embedding time is not influences by the distance metric and the shown embedding times for the standard t-SNE procedure is representative for the embeddings times of all runs.}
    \label{fig:supp:num_channels}
\end{figure*}

\labelshort[S5]{supp:timings} %
\Autoref{fig:supp:neighhorhood_size, fig:supp:num_channels} show the computation time for the distance computation (including feature computation) and subsequent embedding time.
All measurements were conducted on a computer with an Intel i5-9600K processor and a NVIDIA GeForce RTX 2080 SUPER graphics cards.
A corresponding theoretical complexity analysis of each distance is presented in \autoref{sec:method:complex}.
Some measurements show the influence of hardware optimizations  implemented in the used libraries, which influences the computation time, see for example the time behaviour of the Bhattacharyya distance for various neighborhood sizes in \autoref{fig:supp:neighhorhood_size}.

\FloatBarrier
\subsection*{S6: Indian Pines - additional figures} \labelshort[S6]{supp:pines} %

\begin{figure*}[hbt!]
    \begin{center}
        \includegraphics[width=1\linewidth]{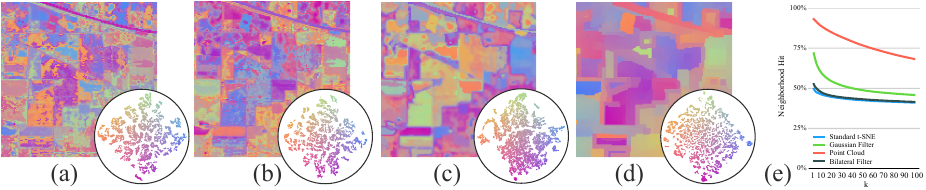}
        \phantomsubcaption\ignorespaces\label{fig:IndianPinesEmbProjection:tsne}
        \phantomsubcaption\ignorespaces\label{fig:IndianPinesEmbProjection:bilinear}
        \phantomsubcaption\ignorespaces\label{fig:IndianPinesEmbProjection:Gauss}
        \phantomsubcaption\ignorespaces\label{fig:IndianPinesEmbProjection:PC}
        \phantomsubcaption\ignorespaces\label{fig:IndianPinesEmbProjection:neighborhood_hit}
        \vspace{-6mm}
    \end{center}
    \caption{\textbf{Overview of the Indian Pines dataset} with different embedding methods. Coloring based on colormapping embedding coordinates, as discussed for \autoref{fig:ArtExamplesOverview}. (a) standard t-SNE, (b) standard t-SNE applied to a bilaterally filtered version of the image, (c) same with a Gaussian filter, (d) our point cloud-based t-SNE, and (e) k-nearest neighbors in embedding space as in \autoref{fig:IndianPines:EmbGtColored:quant} with values for (b) added.}
    \label{fig:IndianPinesEmbProjection}
\end{figure*}

\FloatBarrier

\bibliographystylesupplement{abbrv-doi-hyperref-narrow}
\bibliographysupplement{references}

\end{document}